%% file: main.tex
\pdfoutput=1
\documentclass[10pt,twocolumn,letterpaper]{article}
\usepackage[pagenumbers]{cvpr}
\AddToShipoutPicture{\color[rgb]{.45,.45,.45}\AtTextUpperLeft{\put(0,\LenToUnit{0.8cm}){\parbox{\textwidth}{\centering\cvprtenhv Preprint. Under review.}}}} 
\input{preamble}
\usepackage{booktabs}
\usepackage{array}
\usepackage{placeins}
\definecolor{cvprblue}{rgb}{0.21,0.49,0.74}
\usepackage[pagebackref,breaklinks,colorlinks,allcolors=cvprblue]{hyperref}
\usepackage[capitalize]{cleveref}
\crefname{appendix}{Supp.\ Sec.}{Supp.\ Secs.}\Crefname{appendix}{Supp.\ Sec.}{Supp.\ Secs.}
\def\confName{CVPR}
\def\confYear{2027}

\title{The Platonic Representation Hypothesis in Jacobian-Space:\\Language and Vision Models Agree Approximately Up to a Rotation}
\author{Ross Murphy\\
University College London\\
{\tt\small r.murphy@cs.ucl.ac.uk}
\and
Gabriel Brostow\\
University College London\\
{\tt\small g.brostow@cs.ucl.ac.uk}
}

\begin{document}
\maketitle
\input{sec_cvpr/0_abstract}
\input{sec_cvpr/1_intro}
\input{sec_cvpr/2_related}
\input{sec_cvpr/3_method}
\input{sec_cvpr/4_setup}
\input{sec_cvpr/5_results}
\input{sec_cvpr/6_discussion}
\FloatBarrier
{
    \small
    \bibliographystyle{ieeenat_fullname}
    \bibliography{refs}
}
\clearpage
\maketitlesupplementary
\setcounter{figure}{0}\setcounter{table}{0}\setcounter{equation}{0}
\renewcommand{\thefigure}{S\arabic{figure}}\renewcommand{\theHfigure}{S\arabic{figure}}
\renewcommand{\thetable}{S\arabic{table}}\renewcommand{\theHtable}{S\arabic{table}}
\renewcommand{\theequation}{S\arabic{equation}}\renewcommand{\theHequation}{S\arabic{equation}}
\input{sec_cvpr/7_appendix}
\FloatBarrier
\end{document}

%% file: preamble.tex
\usepackage{cuted} 

\usepackage{currfile} 

\usepackage{microtype}



%% file: sec_cvpr/0_abstract.tex
\begin{strip}
\centering
\includegraphics{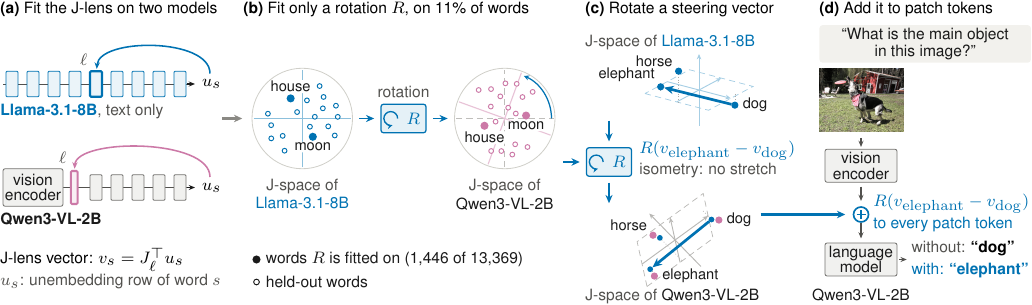}
\captionof{figure}{\textbf{A rotation fitted on a subset of words carries one foundation model's steering vector into another's latent space, changing the object a vision--language model sees in an image.} (a) Pulling word $s$'s unembedding vector $u_s$ back to layer $\ell$ through the context-averaged Jacobian $J_\ell$ defines its J-lens vector $v_s=J_\ell^{\top}u_s$. (b) $R$ is fitted between the two J-spaces on 1,446 of the 13,369 shared single-token words (11\%). (c) As an orthogonal Procrustes map (preserving lengths and angles on the matched subspace without rescaling or shear), $R$ maps text-only Llama-3.1-8B's held-out $v_{\text{elephant}}-v_{\text{dog}}$ into Qwen3-VL-2B's visual J-space, and (d) adding it to visual patch tokens on a dog photo flips the answer from ``dog'' to ``elephant''.}
\label{fig:teaser}
\vspace{0.1em}
\end{strip}

\begin{abstract}
The Platonic Representation Hypothesis (PRH) posits that AI models converge toward a shared representation of the world, even when trained on different data modalities.
Previous work has found evidence for the PRH in the activations of vision and language models.
In this work, we ask whether vision and language models also converge in how they act on these representations: precisely, in how these internal representations affect the models' output.
At a chosen layer of a transformer, the Jacobian lens gives every word in the vocabulary (or class in an image classifier) a vector in the hidden state: the vector that most increases the model's tendency to generate that word, at that position or later, or to predict that class.
Collectively, these vectors make up what is referred to as the model's \emph{J-space}.
We find evidence of the PRH in J-space across vision and language models, and furthermore, we find that J-spaces mostly converge up to a rotation.
That is, a single rotation, fitted on a subset of words, approximately maps one model's J-space onto another's, and so applies to words and classes it was not fitted on.
A word's vector, rotated from a text-only language model's J-space into the visual patch tokens of a vision--language model, changes the object the vision--language model says is in the image (and only in the image region where it is added).
Rotated into an image generator's text conditioning, a word's vector changes the object in the generated image.
Rotated into an ImageNet classifier (trained without text), it raises that object's class score above almost all others, and vice versa.
This suggests that models converge not only in how they represent the world, but also in how they act on it.
\end{abstract}

%% file: sec_cvpr/1_intro.tex
\section{Introduction}
\label{sec:intro}

In Plato's allegory of the cave \citep{plato_cave}, chained occupants who see only shifting shadows on a wall still infer the same figures outside the cave, despite each observing a different projection.
Recent work argues that modern foundation models behave in much the same way: even when trained on distinct data modalities, larger networks converge toward a shared statistical model of reality, a claim known as the Platonic Representation Hypothesis (PRH) \citep{huh2024platonic,balestriero2026proving}.
\citet{huh2024platonic} relate this hypothesis both to convergent realism in the philosophy of science \citep{putnam1982three} and to multiview identifiability in representation learning \citep{tian2020contrastive,zimmermann2021contrastive,richens2024robust}.

In this paper, we ask a distinct but related question: do foundation models also converge on a shared manner of \emph{thinking about} representations?
Put more plainly: we know that models appear to converge on a shared latent \emph{representation} of abstract concepts, but do they also converge on a shared model of how intermediate representations propagate through the network's layers into the generated output?

We investigate this question using the Jacobian lens (J-lens) \citep{gurnee2026verbalizable}, an interpretability technique that identifies an output-facing ``workspace'' inside transformers.
This workspace has parallels with the concept of a ``Global Workspace'' in cognitive science \citep{Baars}: the theory suggests that human brains post intermediate representations to a ``central hub'', from which downstream processes read \citep{dehaene,mashour}, a process which sounds remarkably similar to how transformers route intermediate residual-stream states, through layers, to reach the final output layer.
As formalized in \cref{sec:method}, the J-lens averages, over contexts, the Jacobian of a transformer's final-layer state with respect to an earlier layer $\ell$.
Pulling each token's unembedding row (or class weight) back through this Jacobian yields its \emph{J-lens vector}, the direction at layer $\ell$ that most increases that output's logit to first order, and together these vectors form the model's \emph{J-space}.

We estimate and compare the J-spaces of 45 open-weight language, vision, and multimodal models across scales.
We compare the J-spaces by a measure of their geometries. The measure we use is debiased centered kernel alignment (CKA) \citep{song2012feature,kornblith2019similarity} on centered pairwise cosine matrices. Using this measure, we find that cross-family language-model J-space alignment rises with depth (of layers in the transformer), and with scale, up to ${\sim}1$B parameters before plateauing (\cref{sec:align}).
We also test causal control. We fit a single orthogonal Procrustes map (a pure rotation on the matched subspace preserving lengths and angles without rescaling or shearing) on a subset of words, and then map this J-lens into the space of another model. Across language models, we were surprised to find that rotation recovers 92\% of the steering capability of an unconstrained linear map (\cref{sec:transfer}).

Even more strikingly, this rotational alignment crosses modalities: J-lens vectors from a language model that has only ever read text can also be rotated directly into models that perceive or generate images (\cref{sec:crossmodal}).
Rotated into the visual patch stream of Qwen3-VL \citep{bai2025qwen3vl}, a word's J-lens vector changes the object Qwen3-VL reports seeing in a photo, and only in the region where it is added (\cref{fig:teaser}).
Rotated into the text conditioning of SDXL \citep{podell2024sdxl}, it overrides the prompt and makes the generator draw the new object.

Does this cross-modal compatibility only arise because Qwen3-VL and SDXL were trained with paired images and text?
To test whether vision and language models align in J-space without multimodal supervision, we examine 19 standard ImageNet Vision Transformers (ViTs) \citep{dosovitskiy2021image} trained strictly to classify images into 1,000 categories without ever seeing text.
Pulling those class weights back through the ViT yields a visual J-space where one rotation fitted on a subset of class names lets a language model's J-lens vector steer a ViT's predictions on held-out classes, and lets a ViT's J-lens vector steer a language model's word predictions in reverse (\cref{sec:vit}).

Finally, we investigate why it may be the case that we observe these patterns in the data, and we elucidate our hypotheses and supporting analyses in \cref{sec:why}.

%% file: sec_cvpr/2_related.tex
\section{Related work}
\label{sec:related}

\paragraph{Representational convergence and the Platonic Representation Hypothesis.}
Comparing neural systems via stimulus-by-stimulus similarity matrices began in neuroscience with representational similarity analysis \citep{kriegeskorte2008representational}.
In deep networks, early cross-seed studies showed that while individual neurons rarely match one-to-one, populations span shared subspaces \citep{li2016convergent}, formalized via PWCCA \citep{morcos2018insights} and CKA \citep{kornblith2019similarity}.
\citet{huh2024platonic} synthesized these trends into the Platonic Representation Hypothesis (PRH): as models scale, networks trained across distinct objectives and modalities converge toward a shared statistical model of reality, reflected in rising mutual $k$-NN \citep{huh2024platonic} and CKA \citep{maniparambil2024vision} agreement between vision and language encoders.
Subsequent work sharpened where this holds: contrastive and paired-view JEPA objectives identify latents up to an orthogonal transformation \citep{zimmermann2021contrastive,balestriero2026proving}, whereas cross-entropy predictors identify them up to an invertible linear map \citep{roeder2021linear,reizinger2025cross,balestriero2026proving}.
By and large however, this literature compares models' \emph{static activations}, and the few sensitivity studies that exist are limited to input gradients \citep{wang2024great} and pullbacks \citep{yavari2026beyond}. We shift the focus from input encodings to \emph{output propagation}, and ask whether models converge in how perturbations in their latent spaces steer their outputs.

\paragraph{Latent-space alignment, zero-shot transfer, and model stitching.}
If independently trained models organize concepts in a similar relative shape within different coordinates, how simple can a map between them be?
In cross-lingual NLP, \citet{mikolov2013exploiting} linked word embeddings by a linear map, and \citet{xing2015normalized} constrained it to an orthogonal Procrustes rotation on normalized vectors, which \citet{conneau2018word} extended to unsupervised alignment.
In parallel, zero-shot learning in computer vision bridged image classifiers and word vectors by projecting visual features into word space \citep{frome2013devise,socher2013zero}, combining class probabilities over word embeddings \citep{norouzi2014zero}, or synthesizing unseen classifier weights \citep{changpinyo2016synthesized,xian2019zero}, while model stitching \citep{lenc2015understanding,csiszarik2021similarity,bansal2021revisiting} connected internal layers across networks via learned linear maps.
Building on Procrustes metrics \citep{williams2021generalized}, \citet{moschella2023relative} and \citet{maiorca2023latent} showed that encoders differ mainly by angle-preserving maps, extended to same-architecture seeds \citep{kapoor2025bridging,mccann2026polymorphism} and width-mismatched transformers \citep{theus2025generalized} alongside linear \citep{lahner2024direct}, non-linear \citep{jha2025harnessing}, and stitched \citep{smith2025functional} variants.
Numerous classic zero-shot and recent cross-modal studies \citep{merullo2023linearly,maiorca2023latent,moayeri2023text,schnaus2025blind} map frozen visual features into text or CLIP spaces. Our approach differs: we fit an orthogonal Procrustes map directly between internal J-spaces across different models and modalities, and then test whether rotated J-lens vectors act causally in both models.

\paragraph{Vocabulary lenses and cross-model steering.}
To inspect how transformers form predictions across depth, ``vocabulary lenses'' project intermediate states into token space: the logit lens (a blog post by \citet{nostalgebraist2020interpreting}) applies $W_U$ directly, the tuned lens \citep{belrose2023eliciting} learns an affine correction, and the Jacobian lens (J-lens) \citep{gurnee2026verbalizable} pulls unembedding rows back through the context-averaged Jacobian within single Claude models, with multimodal analogs projecting through CLIP \citep{radford2021learning,gandelsman2024interpreting}, VLM unembeddings \citep{neo2025towards,jiang2025interpreting}, or diffusion conditioning \citep{toker2024diffusion}.
Whereas lenses read states passively, activation steering \citep{turner2023steering,rimsky2024steering,tan2024analyzing} intervenes causally during inference.
Early work extracted steering directions within the target model itself \citep{gan2026textual,baumann2025continuous,gandikota2024concept,gurnee2026verbalizable}, while recent studies transport steering vectors across models via least-squares \citep{huang2025cross,lee2025shared,gan2026beyond}, affine or non-linear maps \citep{oozeer2025activation,chen2025transferring}, or orthogonal SAE maps \citep{puri2026atlas}. Concurrent preprints explore cross-architecture activation transfer \citep{agarwal2026cross,cardenaspiepereit2026architecture}, rotational safety transfer \citep{poppi2026share,bosco2026locating}, and token-indexed J-lens readouts \citep{wang2026jcomm}.
In contrast, we show that one Procrustes rotation fitted on words in J-space transfers held-out J-lens vectors into Qwen3-VL patches, SDXL conditioning, and text-free ImageNet ViTs, steering both bidirectionally.

%% file: sec_cvpr/3_method.tex
\section{Method}
\label{sec:method}

\begin{figure*}[t]
\centering
\includegraphics[width=0.72\textwidth]{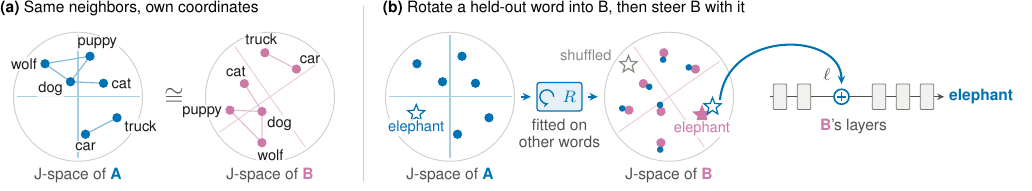}
\caption{\textbf{Two models arrange words in a similar relative shape in their own coordinates, so a rotation fitted on other words carries a held-out word between them.} (a) Models $A$ and $B$ place shared words and angular neighbors in similar relative shapes in distinct coordinate bases (\cref{eq:kj}). (b) An orthogonal Procrustes map $R$ (\cref{eq:rot}), fitted without \emph{elephant}, maps $A$'s held-out vector toward $B$'s own and steers $B$ (\cref{eq:steer}).}
\label{fig:explainer}
\end{figure*}

Our goal is to determine whether distinct neural networks organize their first-order output sensitivities in the same geometric shape, and whether that shared geometry supports causal control across models.
We proceed in three stages: we define each model's output-sensitivity geometry via the Jacobian lens (\cref{sec:method-jspace}), we then compare these geometries observationally across models by comparing kernel alignment metrics (\cref{sec:method-align}), and finally, we test whether an orthogonal rotation fitted on a subset of words in one model can transfer causal steering vectors to held-out concepts in another model (\cref{sec:method-transfer}).

\subsection{Output-Sensitivity Geometry (J-Space)}
\label{sec:method-jspace}

\paragraph{The Jacobian lens.}
Consider a transformer with residual-stream activations $h_{\ell,p}\in\mathbb{R}^d$ at layer $\ell$ and token position $p$, and final-block output $h_{L,p'}\in\mathbb{R}^d$ at position $p'$ \citep{elhage2021mathematical}.
To summarize how intermediate states at layer $\ell$ influence the output, the \emph{Jacobian lens} (J-lens) \citep{gurnee2026verbalizable} replaces all blocks between $\ell$ and $L$ with their first-order linear approximation around real inputs, summing the Jacobian of $h_{L,p'}$ with respect to $h_{\ell,p}$ over readout positions $p'\ge p$ and averaging over source positions $p$ and contexts $x$:
\begin{equation}
  J_\ell \;=\; \mathbb{E}_{x,\,p}\!\left[\sum_{p'\ge p}\frac{\partial h_{L,p'}}{\partial h_{\ell,p}}\right] \in\mathbb{R}^{d\times d}.
  \label{eq:jlens}
\end{equation}

\paragraph{J-lens vectors.}
At the output layer, each token $t$ is scored by an effective readout vector $u_t$, defined as row $t$ of $W_U$ with the linear parts of the final normalization (gain and, for LayerNorm, mean subtraction) folded in (\cref{app:lens}).
Pulling $u_t$ backward through $J_\ell$ yields token $t$'s \emph{J-lens vector} at layer $\ell$,
\begin{equation}
  v_t \;=\; J_\ell^{\top}u_t.
  \label{eq:v}
\end{equation}
Since token $t$'s logit is $z_t = u_t^{\top}h_L/\sigma(h_L)$ up to a constant (\cref{app:lens}), holding $\sigma(h_L)$ fixed under a perturbation $\delta$ at $h_{\ell,p}$ shifts $t$'s unscaled readout to first order by
\begin{equation*}
  \mathbb{E}_{x,p}\!\left[\sum_{p'\ge p}\Delta\!\left(u_t^{\top}h_{L,p'}\right)\right] \;=\; v_t^{\top}\delta + O(\|\delta\|^2).
\end{equation*}
Thus, $\hat v_t = v_t/\|v_t\|$ is the direction at layer $\ell$ that most strongly promotes token $t$, and we call the geometry of $\{v_t\}$ the model's \emph{J-space} (whereas \citet{gurnee2026verbalizable} reserve the term for their nonnegative cone).

\paragraph{Basis invariance.}
Each model expresses its J-space in an arbitrary internal coordinate system.
Applying an invertible change of basis $h_\ell \mapsto M h_\ell$ at layer $\ell$ and absorbing $M^{-1}$ into subsequent blocks leaves the network's function unchanged while transforming $J_\ell$ and $v_t$:
\begin{equation}
  J_\ell \mapsto J_\ell M^{-1}, \quad v_t \mapsto M^{-\top}v_t, \quad v_t^{\top}\delta \mapsto v_t^{\top}\delta.
  \label{eq:basis}
\end{equation}
Only the pairings $v_t^{\top}\delta$ are intrinsic.
Because J-lens vectors act on activations as dual linear functionals \citep{park2024linear}, the angles between $\{v_t\}$ are preserved if and only if $M$ is orthogonal up to scale on $\operatorname{span}(\{v_t\})$ ($M^{-\top}\propto M$ there, \cref{sec:why}), rotating both spaces together.

\subsection{Observational Alignment Across Models}
\label{sec:method-align}

If two models share a J-space geometry, each word should have the same angular neighbors in both, regardless of width or coordinate basis (\cref{fig:explainer}a).
Over a shared vocabulary $\mathcal{C}$, we center $\{v_t\}$ and form each model's \emph{cosine kernel}:
\begin{equation}
  K_J[s,t] \;=\; \cos(\bar v_s,\bar v_t), \qquad
  \bar v_t \;=\; v_t - \frac{1}{|\mathcal{C}|}\sum_{c\in\mathcal{C}} v_c.
  \label{eq:kj}
\end{equation}
We compare kernels across families using \emph{debiased} CKA \citep{song2012feature,kornblith2019similarity}, which removes diagonal finite-sample bias so independent kernels score $0$ in expectation and proportional similarity matrices score $1$ (mutual 10-NN agrees, \cref{app:lmalign}).
For comparison, we build an activation kernel $K_h$ identically from $h_\ell(s)$, the mean last-token state over text snippets ending in $s$, centered and clipped at the 95th percentile \citep{huh2024platonic}, while leaving $K_J$ unclipped.
To compare architectures of different depths, we sample 16 evenly spaced blocks from $0\%$ (first block) to $100\%$ (penultimate, before $J_L = I$), taking $87\%$ ($13$/$15$) as late layers and $53\%$ as mid-depth.


\subsection{Causal Steering and Cross-Model Transfer}
\label{sec:method-transfer}

\paragraph{Behavioral meaning: steering and side effects.}
The angles in J-space have a direct behavioral meaning.
We can steer a language model toward word $s$ by perturbing its residual stream along $\hat v_s$ at all positions $p\ge 16$ of 16 held-out 128-token prompts ($r_\ell = \operatorname{median}\|h_{\ell,p}\|$, $\alpha = 1$):
\begin{equation}
  h_{\ell,p} \;\leftarrow\; h_{\ell,p} + \alpha\, r_\ell\,\hat v_s, \qquad \hat v_s = \frac{v_s}{\|v_s\|}.
  \label{eq:steer}
\end{equation}
Perturbing all $p \ge 16$ accumulates upstream pushes from $p \le p'$ to match \cref{eq:jlens}, while downstream normalization re-projects the stream onto the sphere (\cref{app:steer}).
With typical scale $\sigma$, this shifts every word $t$'s logit to first order:
\begin{equation}
  \Delta z_{st} \;\approx\; \frac{\alpha\, r_\ell}{\sigma}\, v_t^{\top}\hat v_s,
  \label{eq:sidefx}
\end{equation}
proportional to row $s$ of the J-lens Gram matrix divided by $\|v_s\|$, predicting the entire profile of steering side effects.
Since observed log-probability shifts $\Delta_{st} = \Delta\log p_{st} = \Delta z_{st} - \Delta\log Z_s$ carry a per-row softmax offset $\Delta\log Z_s$, we compare \cref{eq:sidefx} via Spearman correlation against the double-centered shifts:
\begin{equation}
  \tilde\Delta_{st} \;=\; \Delta_{st} - \bar\Delta_{s\cdot} - \bar\Delta_{\cdot t} + \bar\Delta_{\cdot\cdot}.
  \label{eq:dc}
\end{equation}

\paragraph{Orthogonal Procrustes transfer.}
To test whether one model's J-lens vectors can causally control another, we split shared concepts into a fitting set $\mathcal{F}$ and a stem-disjoint held-out test set (\cref{fig:explainer}b), stacking $\mathcal{F}$-centered vectors $\tilde v^A_t, \tilde v^B_t$ ($t\in\mathcal{F}$) into $\tilde X_A \in \mathbb{R}^{|\mathcal{F}|\times d_A}$ and $\tilde Y_B \in \mathbb{R}^{|\mathcal{F}|\times d_B}$.
Because widths $d_A, d_B$ differ across models, we use the orthogonal Procrustes map \citep{schonemann1966generalized},
\begin{equation}
  R_{A\to B} \;=\; U V^{\top}, \qquad U\Sigma V^{\top} \;=\; \tilde Y_B^{\top}\tilde X_A,
  \label{eq:rot}
\end{equation}
from the rank-$r$ thin SVD of $\tilde Y_B^{\top}\tilde X_A$ ($r \le \min(d_A, d_B, |\mathcal{F}|-1)$), which projects onto the matched subspace $\operatorname{span}(V)$ and acts there as a pure rotation (allowing reflections) without stretching or shearing axes (\cref{app:transfer,app:why}).

We benchmark this \emph{rotation} against ridge regression (\cref{app:ridge}),
\begin{equation}
  W_{A\to B} \;=\; \arg\min_{W}\sum_{t\in\mathcal{F}}\|W\tilde v^A_t - \tilde v^B_t\|^2 + \lambda\|W\|_F^2,
  \label{eq:ridge}
\end{equation}
hereon referred to as the \emph{best linear map}, alongside $B$'s own vector $v^B_s$, a random direction, a \emph{shuffled rotation} on $(\tilde v^A_t, \tilde v^B_{\pi(t)})$, and activation maps fitted on $h_\ell(s)$.

\paragraph{Modality-specific receivers and classifier readout.}
For a held-out concept $s$ (or swap $s \to s'$), we form $\delta = R\tilde v^A_s$ (or $R(\tilde v^A_{s'} - \tilde v^A_s)$) and steer language models via \cref{eq:steer} with $\delta/\|\delta\|$ and centered norm $\tilde r_\ell = \operatorname{median}\|h_{\ell,p} - \bar h_\ell\|$.
Other models which receive this steering vector, `receivers', adapt only the readout and step size: in \emph{ViTs}, $W_U$ is the classification head, $p$ indexes patches, and $h_L$ is the final class token (\cref{app:vit}). In \emph{Qwen3-VL}, $J_\ell \in \mathbb{R}^{d_{\mathrm{out}}\times d_{\mathrm{in}}}$ pulls answer-predicting decoder readouts back to final-vision-block patches $x_p\in\mathbb{R}^{1024}$ before the patch merger (\cref{app:vision}).
Each receiver is steered along the unit vector $\delta/\|\delta\|$, with a step size matched to its own scale:
\begin{equation}
  \begin{aligned}
  \text{ViT:}\;\; h_{\ell,p} &\leftarrow h_{\ell,p}+r_\ell\,\delta/\|\delta\|,\\
  \text{Qwen3-VL:}\;\; x_{p} &\leftarrow x_{p}+a\,r\,\delta/\|\delta\|,\\
  \text{SDXL:}\;\; e_j &\leftarrow e_j+1.5\,m_j\,\delta_j/\|\delta_j\|.
  \end{aligned}
  \label{eq:recv}
\end{equation}
Here $r_\ell$ and $r$ are the median patch-token norms of the ViT and of Qwen3-VL, $a$ is a strength calibrated on class pairs disjoint from the test classes (\cref{app:vision}), and $e_1,\dots,e_5$ are the five conditioning blocks of SDXL's two CLIP text encoders, each with its own map $\delta_j$ and median inter-class distance $m_j$ (\cref{app:gensteer}).
For classifiers over labels $\mathcal{Y}$, we score steering toward $c$ by its relative logit change (log geometric-mean odds increase):
\begin{equation}
  \Delta_{\mathrm{rel}}(c) \;=\; \mathbb{E}_x\!\left[\Delta z_c - \frac{1}{|\mathcal{Y}|-1}\sum_{c'\neq c}\Delta z_{c'}\right].
  \label{eq:rel}
\end{equation}

%% file: sec_cvpr/4_setup.tex
\section{Experimental setup}
\label{sec:setup}

\paragraph{Models and shared vocabulary.}
We evaluate 23 base language models across five families (160M--32B): Pythia \citep{biderman2023pythia} (160M--12B), Qwen3 \citep{yang2025qwen3} (0.6B--14B), OLMo-2 \citep{teamolmo2024olmo} (1B--32B), Llama-3 \citep{grattafiori2024llama} (1B--8B), and Gemma-3 \citep{gemmateam2025gemma} (1B--27B, multimodal at $\ge$4B), using the 17 models in four matched tiers (${\sim}1$B, $3$B, $8$B, $13$B, 56 pairs, \cref{tab:sizes}) for cross-family transfer.
For vision and multimodal tests, we evaluate 19 supervised ImageNet ViTs \citep{deng2009imagenet} (DeiT \citep{touvron2021training}, DeiT III \citep{touvron2022deit}, AugReg \citep{steiner2022how}, 5.7M--632M), Qwen3-VL (2B, 4B, 8B) \citep{bai2025qwen3vl}, and SDXL \citep{podell2024sdxl}.
Eligible concepts are lower-case, alphabetic, $\ge$3-letter, non-stopword WordNet entries \citep{miller1995wordnet} that appear as single leading-space tokens in all five tokenizers: 13,369 of the 40,000 most frequent English words qualify, from which $\mathcal{C}$ samples 1,456 words across frequency deciles (\cref{app:robust}).
J-lenses use bfloat16 on 256 WikiText-103 \citep{merity2017pointer} 128-token passages (128 at 27B/32B) and 1,000 ImageNet images (within-model steering validation is reported in \cref{app:steer}).

\begin{table}[!ht]
\centering\footnotesize
\setlength{\tabcolsep}{2.2pt}
\resizebox{\columnwidth}{!}{%
\begin{tabular}{@{}llll@{}}
\toprule
Setting & Source $\to$ target depth & Fitted on & Tested on \\
\midrule
LM $\to$ LM & J-lens (53\% or 87\%) & 6,000 words & 300 words \\
LM $\leftrightarrow$ ViT & 87\% $\to$ 73\% (rev.\ 73\% $\to$ 53\%) & 582--588 names & 120 classes \\
Qwen3-VL & LM 53\% $\to$ final vision block & 1,446 words & 12 classes \\
SDXL & LM 87\% $\to$ CLIP text blocks & 6,980 words & 40 / 229 cls. \\
\bottomrule
\end{tabular}}
\caption{\textbf{Each map is fitted on words or class names and tested on stem-disjoint held-out concepts.} ViT fits omit test classes' WordNet neighbors, Qwen3-VL omits $\mathcal{C}$ words sharing a COCO stem (24 pairs, 12 classes), and SDXL uses one word per stem.}
\label{tab:sizes}
\end{table}

%% file: sec_cvpr/5_results.tex
\section{Results}
\label{sec:results}

\subsection{Language models of different families align in J-space beyond static unembeddings}
\label{sec:align}
\label{sec:raw}
\label{sec:size}

Before asking whether one model's J-space can be rotated onto another's, we test a more basic geometric prerequisite: do independently trained models arrange words in the same relative shape at all?
If two models' output-sensitivity geometries agree, then any pair of words that point in similar steering directions in one J-space should point in similar directions in the other, regardless of each model's internal coordinates.
Mathematically, this asks whether their centered J-space cosine kernels $K_J$ (\cref{eq:kj}) are proportional, which debiased CKA measures directly, without requiring us to fit a map between models.

\begin{figure}[!t]
\centering
\includegraphics[width=0.90\linewidth]{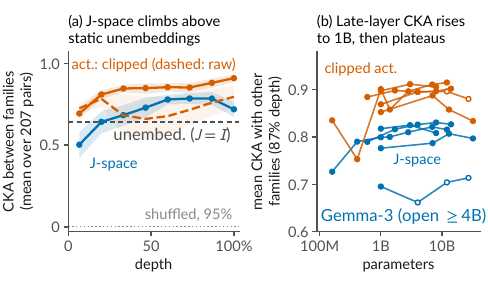}
\caption{\textbf{Cross-family J-space CKA climbs well above static unembeddings ($J=I$) with depth and plateaus above 1B parameters.} Debiased CKA over 1,456 words (\cref{eq:kj}). (a) Mean over 207 cross-family pairs by depth (95\% intervals), with clipped and raw (dashed) activations, static unembeddings $K_U$ ($J=I$), and the 95th-percentile shuffled null. (b) Late-layer (87\%) CKA vs.\ size (open: multimodal Gemma-3 $\ge$4B).}
\label{fig:cka}
\end{figure}

Across 1,456 words shared by 23 language models from five families (207 cross-family pairs), J-space CKA climbs steadily with depth from 0.50 at the first block to 0.78 in late layers (87\% depth), against 0.002 when word labels are shuffled (\cref{fig:cka}a).

A natural question is whether this agreement simply reflects the static token unembedding matrix $W_U$ before any intermediate layers act. Setting $J_\ell = I$ in \cref{eq:v} leaves each word (token) represented solely by its static unembedding vector $u_t$ --- this leaves us with a kernel $K_U$ which is just an unembedding kernel, and this reaches 0.64 CKA.

Thus, we can see that pulling the unembedding matrix back through the network's layers lifts alignment well above the static baseline of 0.64, to 0.78, which actually matches raw activation alignment (0.76, vs.\ 0.88 clipped \citep{huh2024platonic}).

Within model families, late-layer J-space CKA rises by 0.05 per tenfold parameter increase, driven by growth up to ${\sim}1$B parameters before plateauing among larger models. The one notable outlier is Gemma-3, sits slightly lower at 0.66--0.71 (\cref{fig:cka}b, cf.\ \citet{groger2026revisiting}, \cref{app:size}).

\paragraph{Separating J-space from skip connections and word statistics.}
Because residual skip connections pass $W_U$ through even untrained layers ($J_\ell\approx I$), re-initializing the blocks across ten models while keeping each model's trained $W_U$ retains 0.57 CKA on unweighted $K_J$. The learned layers thus lift this by $+0.21$ to 0.78 (\cref{app:decomp}).

Three further checks confirm that this gain reflects learned multi-layer propagation, rather than it simply being the unembedding matrix, or corpus statistics. First, regressing each model's static $K_U$ out of its $K_J$ leaves a Jacobian residual whose CKA climbs with depth from 0.40 to 0.54 (vs.\ 0.002 shuffled). Second, regressing out corpus PMI and 300-dimensional GloVe vectors \citep{pennington2014glove} leaves late-layer $K_J$ CKA at 0.72 (\cref{sec:tables,app:decomp}).
Third, weighting J-lens vectors by hidden-state covariance on disjoint OpenWebText passages yields a basis-invariant kernel $K_S$ (\cref{app:states}) that reaches 0.91 CKA in late layers, whereas holding those states fixed and substituting $J_\ell=I$ or random-block Jacobians drops CKA to 0.58 and 0.54.

\subsection{One rotation carries J-lens steering profiles across language-model families}
\label{sec:transfer}
\label{sec:steer}

Kernel agreement establishes that two J-spaces have a similar relative shape, yet it remains correlational.
If two J-spaces truly differ mainly by an orthogonal change of coordinates, then a single rotation fitted on one set of words should transport a \emph{held-out} word's J-lens vector into another model and causally steer its predictions.
Across 56 directed cross-family pairs (17 models), we fit the Procrustes rotation (\cref{eq:rot}) on 6,000 words and test causal steering on 300 unseen, stem-disjoint words (\cref{tab:sizes}).

\begin{figure}[!t]
\centering
\includegraphics[width=0.88\linewidth]{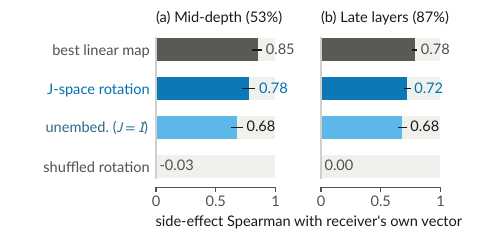}
\caption{\textbf{A J-space rotation beats unconstrained linear maps on static unembeddings ($J=I$) and recovers 92\% of the best linear map.} Spearman correlation of double-centered shifts (\cref{eq:dc}) across 2,499 non-target words with the receiver's own vector at (a) mid-depth (53\%) and (b) late layers (87\%) over 56 pairs and 300 held-out words (95\% intervals).}
\label{fig:lmtransfer}
\end{figure}

The geometric signature of J-space lies in its steering \emph{side effects} (\cref{eq:sidefx}): pushing a model toward \emph{dog} should also raise \emph{puppy} more than \emph{carburetor}, in proportions governed by the receiver's J-lens Gram matrix.
Across all 2,499 non-target candidates, double-centered shifts (\cref{eq:dc}) under a J-space rotation correlate with those under the receiver's own vector at Spearman 0.78 at mid-depth (53\%) and 0.72 in late layers (87\%), recovering 92\% of the best linear map (ridge: 0.85 and 0.78, vs.\ $\le 0.00$ shuffled, \cref{fig:lmtransfer}) and beating even an unconstrained linear map between static unembeddings ($J_\ell=I$: 0.68 at both depths, and 0.44--0.48 vs.\ 0.85--0.86 at 20--33\% depth, \cref{app:transfer}).
Static activations also transfer in text (0.81 vs.\ 0.89 ridge).

\subsection{Text-only vectors steer Qwen3-VL visual patches and SDXL conditioning}
\label{sec:crossmodal}
\label{sec:gensteer}

If J-space reflects a modality-independent structure, a J-lens vector extracted from a model that has \emph{only ever read text} should act coherently inside a model that perceives or generates images.
In Qwen3-VL, visual patch tokens $x_p\in\mathbb{R}^{1024}$ leave the final vision block upstream of the patch merger (\cref{eq:recv}), where no vocabulary unembedding exists and raw vision-block patch activations averaged over captioned images share \emph{no} detectable map-free geometry with language-model activations (Gram correlation 0.07 vs.\ 0.12 for the 99th-percentile null).
Pulling Qwen3-VL's answer readouts back through $J_\ell$ into the vision encoder, however, assigns every word a visual J-lens vector directly from model weights without single-object crops, aligning with language-model J-spaces (Gram correlation 0.33, above the 99th percentile) and steering nearly twice as often as patch activations fitted on the same 512 captioned words (80\% vs.\ 41\%, \cref{fig:qwenclean,app:transfer,app:why}).

\begin{figure}[!t]
\centering
\includegraphics[width=0.84\linewidth]{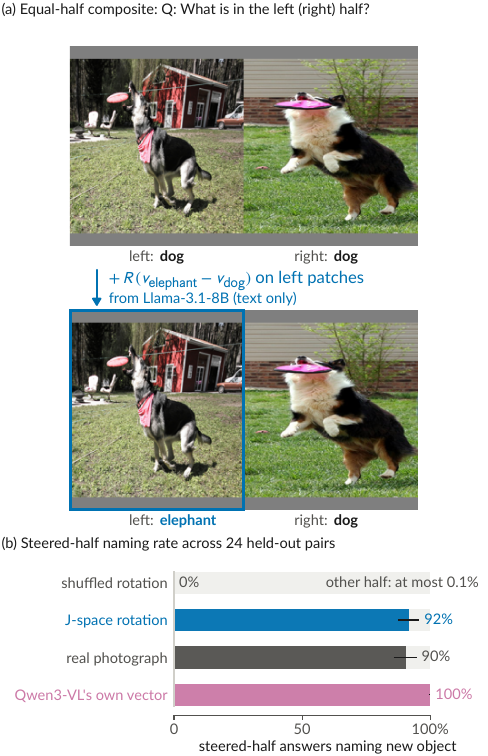}
\caption{\textbf{A text-only J-lens vector rotated into Qwen3-VL steers object perception only on the patches where it is added.} (a) Equal $192\times192$ dog photos on Qwen3-VL-2B's $384\times384$ canvas: adding Llama-3.1-8B's rotated $v_{\text{elephant}}-v_{\text{dog}}$ to left patches flips only the left answer. (b) Steered-half naming rate over four LMs and 24 held-out pairs (unsteered $\le 0.1\%$).}
\label{fig:demo}
\end{figure}

We fit Procrustes rotations on 1,446 non-test words (\cref{tab:sizes}) from four language models (text-only Llama-3.1-8B, Pythia-12B, and OLMo-2-13B, plus multimodal Gemma-3-12B) into Qwen3-VL-2B's visual J-space and transport $v_{\text{new}}-v_{\text{old}}$.
Crucially, the rotated J-lens vector acts spatially wherever it is placed: adding it to one half of a two-dog composite (\cref{fig:demo}a) makes Qwen3-VL-2B report the new object in that half alone on 92\% of tests (vs.\ 100\% for Qwen3-VL's own vector, 90\% for a real photo, 0.05\% unsteered, and 0\% shuffled, \cref{fig:demo}b).
On full images across all four sources, J-space rotations reach 87\% at their calibrated strength ($a=0.08$, 88\% of the own vector at 99\%, and 91\% across text-only sources), matching calibrated ridge ($a=0.05$: 84\%, whereas at matched $a=0.05$ ridge leads 85\% to 72\% before degrading at $a=0.08$, \cref{fig:qwencontrols}).
Both full-rank (87\%) and rank-300 (74\%) J-space rotations far beat 300-dimensional GloVe rotations (51\%), and across Qwen3-VL-2B, 4B, and 8B receivers J-space rotations (84\%, 77\%, and 77\%) beat static unembeddings ($J_\ell=I$: 78\%, 71\%, and 71\%) by ${\sim}6$ points at every scale (\cref{app:vision}).

\paragraph{Visual interaction controls.}
Two checks confirm that rotated J-lens vectors interact with visual perception rather than blindly overwriting the output: whereas $v_{\text{new}}$ alone acts like a strong prior (72\% on old-class photos), $v_{\text{new}}-v_{\text{old}}$ drops to 38\%, 26\%, and 32\% on unrelated, gray, and zeroed images (preserving true names on 49--54\% of unrelated photos), and on a 10-question class-free attribute VQA probe it shifts answers to match the new object's visual attributes on 74\% of questions vs.\ 6\% unsteered and 17--21\% for text prefixes (\cref{app:vision}).

\begin{figure}[!t]
\centering
\includegraphics[width=0.84\linewidth]{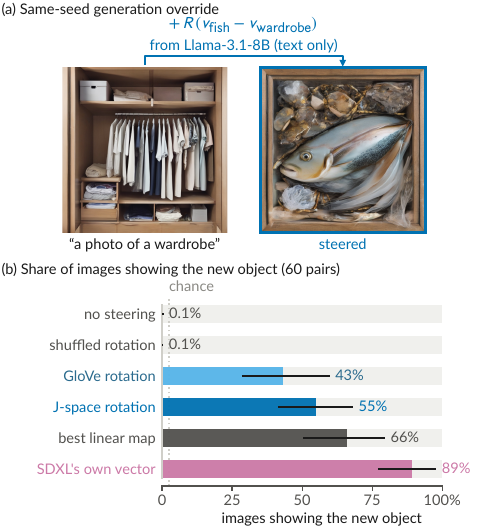}
\caption{\textbf{Translating text-only J-lens vectors into SDXL's CLIP conditioning overrides generated objects and beats GloVe rotations.} (a) Prompted with \emph{a photo of a wardrobe}, SDXL generates a fish under Llama-3.1-8B's rotated $v_{\text{fish}}-v_{\text{wardrobe}}$ (same seed). (b) DINOv2 40-way target rate over four text-only LMs and 60 pairs (chance 1/40).}
\label{fig:gensteer}
\end{figure}

The same alignment extends to image synthesis through SDXL's text conditioning, where five Procrustes maps per text-only language model translate J-lens vectors into the five penultimate blocks of SDXL's two CLIP text encoders on \emph{a photo of a} + word prompts (\cref{eq:recv}), overriding the prompt under $v_{\text{new}}-v_{\text{old}}$ (\cref{fig:gensteer}a).
Across 60 pairs of 40 held-out Open Images \citep{kuznetsova2020open} classes, J-space rotations from four text-only language models make SDXL draw the new object in 55\% of images (DINOv2 \citep{oquab2024dinov2} nearest-centroid), reaching 62\% of SDXL's own vector (89\%) and 83\% of the best linear map (66\%), beating GloVe rotations (43\%, \cref{fig:gensteer}b) and matching activation rotations (51\%, controls $\le 0.2\%$).
On 229 further unseen classes (343 pairs), J-space rotations retain 81\% of ridge (26.3\% vs.\ 32.4\%, \cref{app:gensteer}).

\subsection{Text-free ImageNet ViTs and language models steer each other}
\label{sec:vit}
\label{sec:vitsteer}

Both Qwen3-VL and SDXL were trained with paired text, so this leaves open the question of whether alignment requires multimodal supervision.
Supervised ImageNet ViTs eliminate this confound: they never see text, mapping pixels directly to 1,000 class indices.
Unlike classic zero-shot methods that regress frozen visual features onto word vectors or synthesize classifiers externally \citep{frome2013devise,socher2013zero,norouzi2014zero,changpinyo2016synthesized}, pulling each ViT's classification head back through $J_\ell$ yields an internal visual J-space (\cref{sec:method}) whose cosine kernels across 19 ViTs agree with all 23 language models over 161 single-token class names well above shuffled nulls (\cref{app:vit}).

\begin{figure}[!t]
\centering
\includegraphics[width=0.80\linewidth]{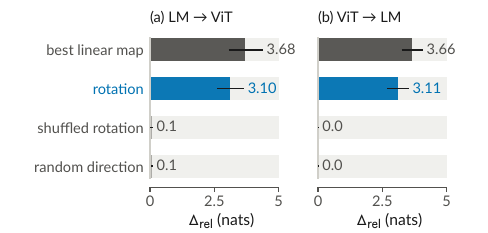}
\caption{\textbf{One class-agnostic rotation per direction lets language models and text-free ImageNet ViTs steer each other, matching 84--85\% of the best linear map.} Relative logit change $\Delta_{\mathrm{rel}}$ (\cref{eq:rel}) in J-space across 437 pairs and 120 held-out classes.}
\label{fig:vit}
\end{figure}

More strikingly, they steer one another bidirectionally through one class-agnostic rotation per direction fitted on stem- and WordNet-disjoint class names (\cref{tab:sizes,fig:vit}).
Adding a language model's rotated J-lens vector to every patch of 500 ImageNet validation images raises the target's $\Delta_{\mathrm{rel}}$ (\cref{eq:rel}) by 3.10 nats (+3.04 net of 0.06 shuffled, ${\sim}20\times$ odds, \cref{fig:vit}a), and rotating a ViT's J-lens vector into a language model raises the class name by 3.11 nats (+3.12 net of $-0.01$ shuffled, \cref{fig:vit}b).
Every one of the 437 ViT--LM pairs transfers positively in both directions, recovering 84--85\% of ridge (3.68 and 3.66 nats).

Against the receiver's own vector (13.5 and 15.7 nats), both cross-modal maps are partial in magnitude: J-space rotation reaches 23\% and 20\% (ridge: 27\% and 23\%, while ViT class-token activations reach 26\% and 44\%, \cref{app:transfer,app:vitsteer}), showing that while text-free cross-modal alignment is partial, nearly all linearly transferable J-space signal is rotational.
Randomizing ViT blocks cuts J-space steering by $3.5\times$ (6.39 to 1.81 nats), and while two-thirds of the logit rise is shared across WordNet siblings, class-specific signal (+0.96 nats) tops all siblings in 47\% of tests vs.\ 29\% chance.

\paragraph{Boundary condition: end-to-end semantic readouts.}
In contrast, vision backbones \emph{without} an end-to-end semantic head align in static activations, but do not appear to align in J-space (\cref{sec:vision,app:dino}): self-supervised DINOv2/v3 ViTs with linear probes trail random-block baselines (0.140 vs.\ 0.174 mutual 10-NN) and decline with scale. Contrastive CLIP/SigLIP2 vision towers lose J-space alignment once their text tower is regressed out, showing that J-space appears to require an end-to-end semantic readout.

%% file: sec_cvpr/6_discussion.tex
\section{Why might a rotation suffice?}
\label{sec:why}

An orthogonal Procrustes map recovers 81--92\% of the best linear map's steering across language models, SDXL, and ViTs, and matches calibrated ridge in Qwen3-VL, even where both maps reach only part of the receiver's own vector.

\paragraph{A rotation exists exactly when steering responses agree.}
Cross-entropy objectives identify representations only up to an invertible linear map $M\in\mathrm{GL}(d)$ \citep{roeder2021linear,reizinger2025cross}, so a rotation is not guaranteed.
By \cref{eq:sidefx}, two models have the same centered side-effect profiles $\tilde v_t^{\top}\hat{\tilde v}_s$ and relative norms $\|\tilde v_s\|$ if and only if their centered Gram matrices agree up to scale, $\tilde Y_B\tilde Y_B^{\top}=c^2\tilde X_A\tilde X_A^{\top}$, which holds if and only if their J-lens vectors differ by a rotation on the matched subspace (\cref{app:why}).
A rotation is therefore as good as the two models' steering responses are alike, and that agreement is approximate: late-layer Procrustes similarity is $s=0.77$ vs.\ 0.40 shuffled, held-out Gram correlations span 0.24--0.77 (0.76 across LMs), and own-vector side effects agree at Spearman 0.44--0.45 in J-space vs.\ 0.62--0.68 in activations (\cref{fig:gram,fig:geomrecv}).

\paragraph{Normalization fixes each model's angles.}
Without normalization, a residual stream that blocks read and write linearly can be re-expressed in any basis $M\in\mathrm{GL}(d)$ without changing the function \citep{elhage2023privileged}, so angles between J-lens vectors would be arbitrary.
The language models and released ViTs we study read the stream through a normalization, $n(h)=\sqrt d\,h/\|h\|$ for RMSNorm, with gains folded into the adjacent linear read \citep{ashkboos2024slicegpt}.
A linear read $L$ can then absorb $h\mapsto Mh$ ($n(Mh)=L\,n(h)$ for all $h\neq 0$) if and only if $M=cQ$ with $Q$ orthogonal, while LayerNorm adds only a shift $M=cQ+\mathbf{1}a^{\top}$ that leaves J-lens vectors unchanged because $v_t\in\mathbf{1}^{\perp}$ (\cref{app:why}).
Each model's J-space angles are thus fixed by its function, so cosine kernels and rotations are well-defined comparisons.

\paragraph{What the best linear map adds.}
Across language models the rotation trails ridge (Spearman 0.72 vs.\ 0.78 in late layers, 0.78 vs.\ 0.85 at mid-depth).
Removing the top 10 outlier directions \citep{kovaleva2021bert,timkey2021all} hurts slightly, whereas a whiten--rotate--recolor map $C_B^{1/2}RC_A^{-1/2}$ that uses \emph{only each model's own covariance} \citep{artetxe2018generalizing} reaches 0.77 and 0.84 (\cref{app:transfer,fig:why}a--c), so ridge's gain is per-axis rescaling, not shear.

\paragraph{A conjecture for what remains.}
Normalization constrains one model at a time, and replacing LayerNorm with Dynamic Tanh in ViT-S does not lower the rotation's share of ridge (\cref{app:dyt}), so it cannot by itself place two independently trained models a rotation apart.
The missing step may be the one the PRH supplies: if models trained on different projections of one world recover the same relations among concepts, as the prisoners of the cave infer the same figures from different shadows \citep{plato_cave,huh2024platonic}, their Gram matrices agree, and by the equivalence above a rotation is the only way they can differ.
We leave tests of this account to future work.

\section{Conclusion}
\label{sec:discussion}

Models trained end-to-end into a discrete or semantic readout (language models, Qwen3-VL at 2B--8B, and text-free ImageNet ViTs) agree in first-order output sensitivity as well as in activations.
Across language-model families, J-space kernels align at CKA 0.78, above static unembeddings (0.64), and one rotation fitted on a subset of words recovers 81--92\% of the steering that the best linear map transfers to held-out concepts.
This extends the Platonic Representation Hypothesis from what models represent to how perturbations of those representations move their outputs.

\paragraph{Scope and limitations.}
Both maps are partial against the receiver's own vector: text-free ViT--LM transfer reaches 20--23\% of it, with two-thirds shared across WordNet siblings.
SDXL is steered through its CLIP text conditioning, and backbones without an end-to-end semantic head align in activations, not J-space (\cref{sec:vit}).
The J-lens is a first-order context average, and our main tests use base models and single-token words (\cref{app:dyt,app:persona,app:generation}) on each map's fitting span.

%% file: sec_cvpr/7_appendix.tex

This supplement has three parts.
First, the \hyperref[app:prelim]{Preliminaries} provides further details and intuition around how the J-lens is computed. 
Second, we show that \hyperref[app:videooverview]{J-space alignment extends to video models}, a result which was left out of the main paper for brevity.
Third, Secs.~A--U give the supplementary material referenced in the main paper. 



\section*{Preliminaries: how the J-lens is computed}
\phantomsection\label{app:prelim}

The J-lens (main \cref{eq:jlens,eq:v}) assigns each token $t$ the vector $v_t = J_\ell^{\top}u_t$, where $J_\ell$ is the Jacobian of the final residual state with respect to layer $\ell$, averaged over contexts and positions, and $u_t$ is token $t$'s effective readout vector.

Understanding two steps of its computation are crucial to understanding the subsequent sections.
First: folding the linear parts of the final normalization layer into the unembedding matrix, and second: averaging Jacobians across many text contexts.

\begin{figure}[t]
\centering
\includegraphics[width=\linewidth]{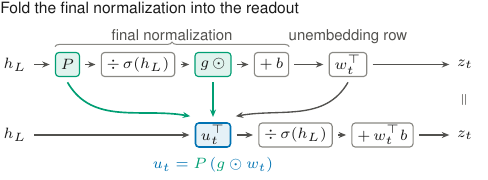}
\caption{\textbf{The final normalization's linear parts fold into the unembedding without changing any logit.} Mean subtraction $P$ and gain $g$ (green) move onto the unembedding row $w_t$ and give the effective readout vector $u_t = P(g\odot w_t)$, which leaves the logit $z_t$ unchanged (\cref{eq:normdef,eq:fold}).}
\label{fig:jlensfold}
\end{figure}

\begin{figure}[t]
\centering
\includegraphics[width=\linewidth]{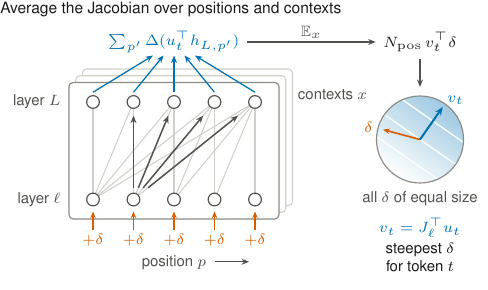}
\caption{\textbf{One averaged vector per token, $v_t = J_\ell^{\top}u_t$, is the steepest first-order direction for steering token $t$.} A perturbation $\delta$ added at every position $p$ of layer $\ell$ reaches only final positions $p'\ge p$ (grey lines; dark: one source position); summing the readout $u_t$ over $p'$ and averaging over contexts $x$ gives $N_{\mathrm{pos}}v_t^{\top}\delta$ (\cref{eq:avg-exact}), so among perturbations of fixed size (disc, shaded by this gain) $\delta \propto v_t$ raises token $t$ most.}
\label{fig:jlensavg}
\end{figure}

Below, we first show how to isolate $u_t$ (\cref{fig:jlensfold}) and then show, by linearity of expectation, that the averaged vector $v_t$ remains the exact steepest first-order steering direction across contexts (\cref{fig:jlensavg}).

\subsection*{Folding final normalization into the unembedding}
\phantomsection\label{app:lens-fold}

Placing $h_L$ immediately before the final normalization layer \citep{gurnee2026verbalizable}, both LayerNorm and RMSNorm map $h_L \in \mathbb{R}^d$ via
\begin{equation}
  \mathrm{norm}(h_L) \;=\; \frac{g \odot P h_L}{\sigma(h_L)} + b,
  \label{eq:normdef}
\end{equation}
with learned gain $g$, bias $b$, scalar norm $\sigma(h_L) > 0$, and mean-subtraction projector $P = I - \tfrac{1}{d}\mathbf{1}\mathbf{1}^{\top}$ (with $P = I, b = 0$ for RMSNorm).
Because $P$ and elementwise multiplication by $g$ are self-adjoint, moving them onto row $w_t$ of $W_U$ defines the \emph{effective readout vector}
\begin{equation}
  u_t \;=\; P(g \odot w_t), \qquad z_t \;=\; \frac{u_t^{\top}h_L}{\sigma(h_L)} + w_t^{\top}b.
  \label{eq:fold}
\end{equation}
Dropping the constant bias $w_t^{\top}b$ and holding $\sigma(h_L)$ fixed to first order leaves token $t$'s logit proportional to $u_t^{\top}h_L$.

\subsection*{Why context averaging is exact to first order}
\phantomsection\label{app:lens-avg}

Adding a fixed perturbation $\delta$ simultaneously to every token position $p$ at layer $\ell$ (\cref{eq:steer}) accumulates upstream pushes from all $p \le p'$ into each readout position $p'$.
Exchanging the causal double sum ($\sum_{p'}\sum_{p \le p'} = \sum_p \sum_{p' \ge p}$) and taking the expectation over contexts $x$ pulls $\delta$ outside both sum and expectation:
\begin{equation}
  \begin{aligned}
    &\mathbb{E}_{x}\!\left[\sum_{p'} \Delta(u_t^{\top}h_{L,p'})\right] \\
    &\qquad \approx \mathbb{E}_{x}\!\left[\sum_{p}\sum_{p' \ge p} u_t^{\top}\frac{\partial h_{L,p'}}{\partial h_{\ell,p}}\,\delta\right]
    = N_{\mathrm{pos}}\, v_t^{\top}\delta,
  \end{aligned}
  \label{eq:avg-exact}
\end{equation}
where $N_{\mathrm{pos}}$ is the number of perturbed positions and $v_t = J_\ell^{\top}u_t$ (\cref{eq:jlens,eq:v}).
Even when single-prompt Jacobians vary, $\hat v_t = v_t/\|v_t\|$ is therefore the exact steepest expected first-order ascent direction for token $t$ (\cref{app:steer}).

\section*{J-space alignment extends to video models}
\phantomsection\label{app:videooverview}
\setcounter{subsection}{0}
\renewcommand{\thesubsection}{V.\arabic{subsection}}\renewcommand{\theHsubsection}{V.\arabic{subsection}}
\newcounter{savedfigure}\setcounter{savedfigure}{\value{figure}}\setcounter{figure}{0}
\newcounter{savedequation}\setcounter{savedequation}{\value{equation}}\setcounter{equation}{0}
\renewcommand{\thefigure}{V\arabic{figure}}\renewcommand{\theHfigure}{V\arabic{figure}}
\renewcommand{\theequation}{V\arabic{equation}}\renewcommand{\theHequation}{V\arabic{equation}}

Video models, which have never been trained on text, agree in J-space up to a rotation, just as language models do. 

\begin{figure*}[t]
\centering
\includegraphics[width=\linewidth]{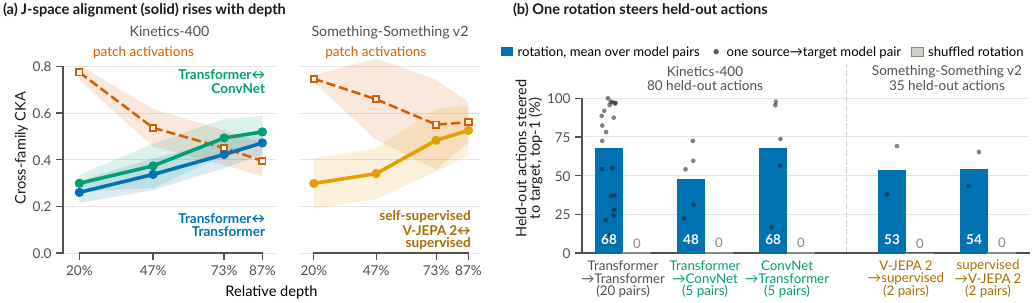}
\caption{\textbf{Video classifiers of different architectures, including self-supervised V-JEPA~2, agree in J-space and steer one another through a single rotation.} (a)~Debiased CKA between video models of different families across relative depth, in J-space (solid) and in patch activations (dashed), for Transformer--Transformer and Transformer--3D ConvNet pairs on Kinetics-400 and V-JEPA~2--supervised pairs on SSv2; shading is the 95\% bootstrap interval over pairs (Kinetics-400) or the range over the two pairs (SSv2). (b)~Top-1 rate: the percentage of clips, averaged over the held-out target actions (80 on Kinetics-400, 35 on SSv2), that the target model assigns to the intended action after steering with a rotated J-lens vector from the source model (chance 0.25\% and 0.57\%); the rotation is fitted on the other actions (318 on Kinetics-400, 139 on SSv2). Each dot is one directed pair (one source model whose rotated J-lens vectors steer one target model). Each blue bar is the mean over the directed pairs in its group; each grey bar is the same mean for a rotation fitted on shuffled action pairs. Details in \crefrange{app:video-method}{app:video-vjepa}.}
\label{fig:videoteaser}
\end{figure*}

We show this holds across multiple architectures, including Transformers, a pure 3D ConvNet (R(2+1)D-18, a convolution-only video network; \cref{app:video-method}), and a self-supervised (V-JEPA style) encoder. 
(\cref{fig:videoteaser}).

We test this on the task of video action recognition. A classification model takes a video as input, and maps the video to one of the 400 human actions of Kinetics-400 \citep{kay2017kinetics} (\eg ``playing guitar'') or one of the 174 object manipulations of Something-Something v2 (SSv2) \citep{goyal2017something} (\eg ``pushing something from left to right'').
Similar to a pure language model, such a classifier ends in a linear head with one row $w_c$ per action $c$.
The classifier model is trained to predict a class index, one of 400 or 174 integers, and never sees the action's name or any other text. We attach the action names only afterwards, to match each action to the same word in language models.

We use the J-lens to `pull' this row back to a hidden layer, $v_c = J_\ell^{\top}w_c$, exactly as $v_t = J_\ell^{\top}u_t$ pulls a word's effective readout vector back in a language model. Here the hidden layer is a grid of patch vectors over space and time, and $J_\ell$ averages the Jacobian over patches and clips (\cref{eq:video-jlens}). Two results follow.

First, J-space alignment between video models of different families rises with depth. This holds both between model families that share an architecture (e.g. VideoMAE and TimeSformer, both Transformers), and across architectures (Transformer--3D ConvNet pairs). Patch-activation alignment, by contrast, falls with depth, and on Kinetics-400 J-space overtakes it in late layers (\cref{fig:videoteaser}a).

Second, one rotation steers held-out actions between architectures, and in both directions: the target model's top prediction becomes the intended action on 48--68\% of clips, against at most 0.2\% for a rotation fitted on shuffled action pairs (\cref{fig:videoteaser}b).
Both results are observed on both Kinetics-400, and on SSv2, where frame order rather than static appearance defines the action. The results also hold for V-JEPA~2-L \citep{assran2025vjepa2}, which is surprising, given V-JEPA~2's encoder is trained only by self-supervision, to predict hidden parts of a video.
For the SSv2 dataset, the authors of V-JEPA~2 kept the encoder's weights unchanged and trained a small classifier head on top: four attention layers that pool the encoder's patch tokens into one vector, then a linear layer over the 174 actions.
The authors released the encoder and classifier together as one SSv2 model, which we use as released, training nothing. The J-lens pulls each action's row back through \emph{V-JEPA~2's classifier} into the encoder.
The four subsections below give the details: \cref{app:video-method} defines the J-lens for video classifiers and lists the models, \cref{app:video-cnn} gives the Kinetics-400 results, including the 3D ConvNet, and \cref{app:video-temp,app:video-vjepa} give the SSv2 and V-JEPA~2 results.

\begin{figure*}[t]
\centering
\includegraphics[width=\linewidth]{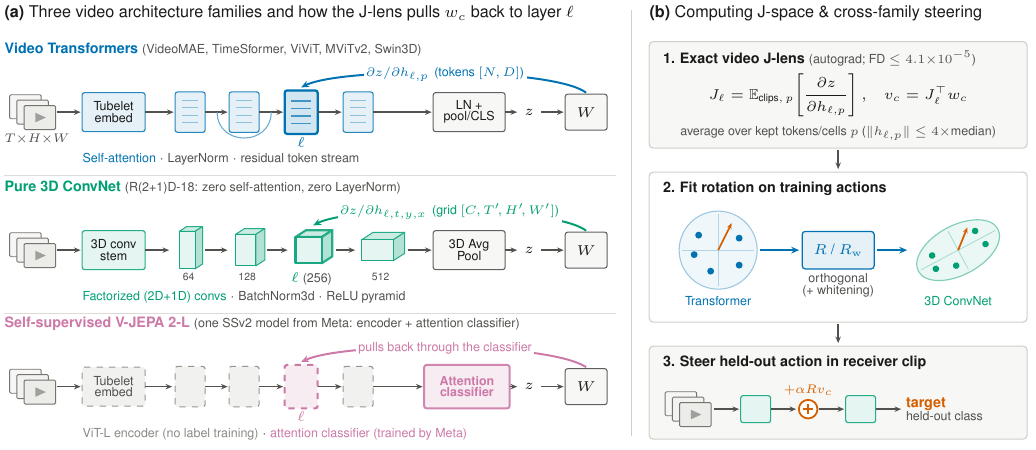}
\caption{\textbf{Exact video J-lenses pull classification-head sensitivities across space and time into three distinct backbones.} (a)~Gradient pullback $\partial z/\partial h_{\ell,p}$ (where $z=h_L(X)$) across Video Transformers, the pure 3D ConvNet R(2+1)D-18 (convolutions and ReLU), and self-supervised V-JEPA~2-L (through V-JEPA~2's classifier: four attention layers and a linear layer). (b)~Spatiotemporal J-lens computation, rotation fitting, and live clip steering.}
\label{fig:videoarch}
\end{figure*}

\subsection{The spatiotemporal J-lens and its form in each architecture}
\label{app:video-method}

\paragraph{Definition.}
Consider any video classifier mapping a $T$-frame clip $X \in \mathbb{R}^{3\times T\times H\times W}$ to a pooled clip representation $h_L(X) \in \mathbb{R}^{d_{\mathrm{out}}}$ entering a linear classification head ($\mathrm{logit}_c = w_c^{\top}h_L(X) + b_c$).
At relative depth $\ell$, the backbone exposes a spatiotemporal grid of feature vectors $h_{\ell,p} \in \mathbb{R}^{d_\ell}$ indexed by position $p = (t, i, j) \in \mathcal{P}_\ell$ across time $t$ and spatial coordinates $(i, j)$ (excluding high-norm outlier positions with $\|h_{\ell,p}\| > 4 \times \mathrm{median}$).

Unlike an autoregressive language model---where position $p$ affects only causal future readouts $p' \ge p$---a video classifier pools the entire spatiotemporal volume into one global feature $h_L(X)$.
We therefore differentiate $h_L(X)$ with respect to every kept position $p \in \mathcal{P}_\ell$ and average uniformly over space, time, and clips (\cref{fig:videoarch}):
\begin{equation}
  J_\ell \;=\; \mathbb{E}_{X}\!\left[\frac{1}{|\mathcal{P}_\ell|}\sum_{p \in \mathcal{P}_\ell} \frac{\partial h_L(X)}{\partial h_{\ell,p}}\right] \in \mathbb{R}^{d_{\mathrm{out}}\times d_\ell},
  \label{eq:video-jlens}
\end{equation}
yielding class sensitivity vectors $v_c = J_\ell^{\top}w_c \in \mathbb{R}^{d_\ell}$.
Although \cref{eq:video-jlens} averages across $p$, each per-position derivative $\partial h_L / \partial h_{\ell,p}$ pulls backward through every downstream temporal layer.
By linearity of differentiation, adding one steering vector $\delta \in \mathbb{R}^{d_\ell}$ simultaneously to every kept position ($h_{\ell,p} \mapsto h_{\ell,p} + \delta$) changes class $c$'s expected logit to first order by
\begin{equation}
  \begin{aligned}
    \mathbb{E}_{X}[\Delta\,\mathrm{logit}_c]
    &= \mathbb{E}_{X}\!\left[\sum_{p\in\mathcal{P}_\ell} w_c^{\top}\frac{\partial h_L(X)}{\partial h_{\ell,p}}\,\delta\right] \\
    &= |\mathcal{P}_\ell|\,v_c^{\top}\delta,
  \end{aligned}
  \label{eq:video-steer}
\end{equation}
making $\hat{v}_c = v_c / \|v_c\|$ the exact steepest expected first-order ascent direction when steering a live video clip.

\paragraph{The three backbone families.}
Three distinct backbone families plug directly into \cref{eq:video-jlens}:
\begin{itemize}
  \setlength{\itemsep}{1pt}\setlength{\parskip}{0pt}\setlength{\parsep}{0pt}
  \item \emph{Video Transformers} (VideoMAE-S/B/L \citep{tong2022videomae}, TimeSformer-B \citep{bertasius2021spacetime}, ViViT-B \citep{arnab2021vivit}, MViTv2-S \citep{li2022mvitv2}, Swin3D-B \citep{liu2022video}): $p = (t, i, j)$ indexes tubelet or frame-patch tokens (excluding any class token), pulling through temporal attention.
  \item \emph{Pure 3D ConvNet} (\texttt{R(2+1)D-18} \citep{tran2018closer}, an 18-layer residual network built only from convolutions, with no attention and no LayerNorm; each 3D convolution over time and space is split into a 2D spatial convolution followed by a 1D temporal one, hence ``(2+1)D''): $p = (t, i, j)$ indexes each cell's channel vector $h_{\ell,p} \in \mathbb{R}^{C_\ell}$ in the 4D feature volume, pulling through 1D temporal convolutions and pooling.
  \item \emph{Self-supervised encoder with a trained classifier} (V-JEPA~2-L \citep{assran2025vjepa2}): $p = (t, i, j)$ indexes the encoder's spatiotemporal tokens, pulling back through V-JEPA~2's classifier (four attention layers and a linear layer) into the encoder.
\end{itemize}

\begin{figure}[t]
\centering
\includegraphics[width=\linewidth]{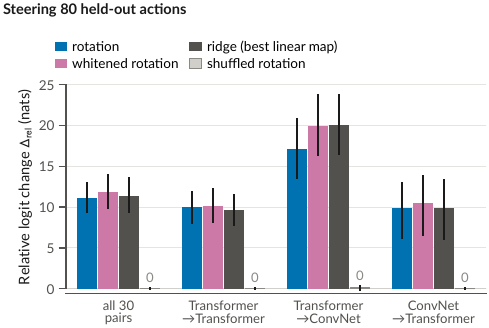}
\caption{\textbf{Six Kinetics-400 families that never see text, including a pure 3D ConvNet, steer 80 held-out actions through one rotation as well as through the best linear map.} Held-out steering, measured as the target action's relative logit change $\Delta_{\mathrm{rel}}$ (main \cref{eq:rel}; crossed-bootstrap 95\% intervals), over all 30 directed pairs and by source$\to$target architecture, for rotation, whitened rotation (\cref{eq:white}), ridge regression (the best unconstrained linear map between two J-spaces, \cref{app:ridge}), and shuffled rotation (grey). Top-1 rates are in \cref{fig:videoteaser}b and alignment across depth in \cref{fig:videoteaser}a.}
\label{fig:videok400}
\end{figure}

\paragraph{Benchmarks and verification.}
Every model is trained only on video and class indices, and matches published single-clip accuracy across Kinetics-400 (8 models, 6 families, 398 classes) and SSv2 (3 models, 174 classes).
Autograd $J_\ell$ matches float64 central finite differences within $4.1 \times 10^{-5}$ relative error across all seven families and four depths (20\%, 47\%, 73\%, 87\%), with late-layer split-half CKA $\ge 0.97$.
Causal steering injects mapped unit directions at the median spatiotemporal patch norm at 87\% depth on held-out classes (318 fit / 80 test on K400; 139 fit / 35 test on SSv2).

\subsection{Pure 3D ConvNets and video Transformers share one causal J-space}
\label{app:video-cnn}

Rotational J-space agreement is not specific to Transformers: the pure 3D ConvNet \texttt{R(2+1)D-18} aligns at least as strongly with all five video Transformer families as they align with one another, and a single orthogonal rotation transfers held-out actions causally in both directions (\cref{fig:videok400}).

\paragraph{Cross-family alignment and causal steering.}
Across depth on Kinetics-400, cross-family J-space CKA climbs steadily for both Transformer--Transformer and Transformer--ConvNet pairs, exceeding late patch activations on every pair (\cref{fig:videoteaser}a).
Causally, one orthogonal rotation fitted on 318 actions transfers all 80 held-out actions across all 30 directed cross-family pairs, recovering 98\% of the steering effect of ridge regression, the best unconstrained linear map between two J-spaces (\cref{app:ridge}; \cref{fig:videok400}).
Among Transformer pairs, plain rotation edges out ridge, and steering out of \texttt{R(2+1)D-18} reaches 57--98\% top-1 into four of the five Transformers; VideoMAE is the hardest target for every source (12--37\%).

\begin{figure}[t]
\centering
\includegraphics[width=\linewidth]{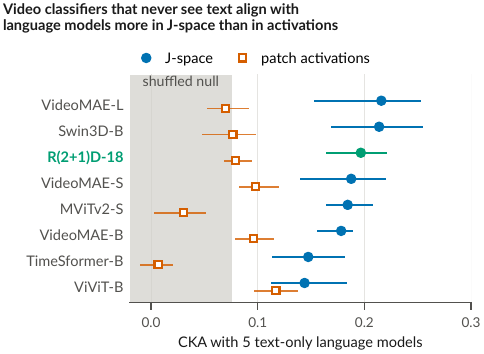}
\caption{\textbf{Video classifiers that never see text align with language models more in J-space than in patch activations.} The video models are trained only on class indices, never on text. Debiased CKA of each of the eight Kinetics-400 models with five text-only language models over 40 shared actions (dot: mean; whiskers: range across the five), in J-space and in patch activations; grey band: shuffled null (largest per-pair 95th percentile).}
\label{fig:videomech}
\end{figure}

\paragraph{Video classifiers that never see text align with text-only language models.}
Extending \cref{sec:vit} to video, all eight Kinetics-400 models, trained only on class indices and never on text (including \texttt{R(2+1)D-18}), align in J-space with five text-only language models across 40 shared action concepts at $2.6\times$ patch-activation alignment and above the shuffled null on all 40 pairs (\cref{fig:videomech}).

\subsection{Rotational J-space holds when temporal order defines the action}
\label{app:video-temp}

\begin{figure*}[t]
\centering
\includegraphics[width=0.96\linewidth]{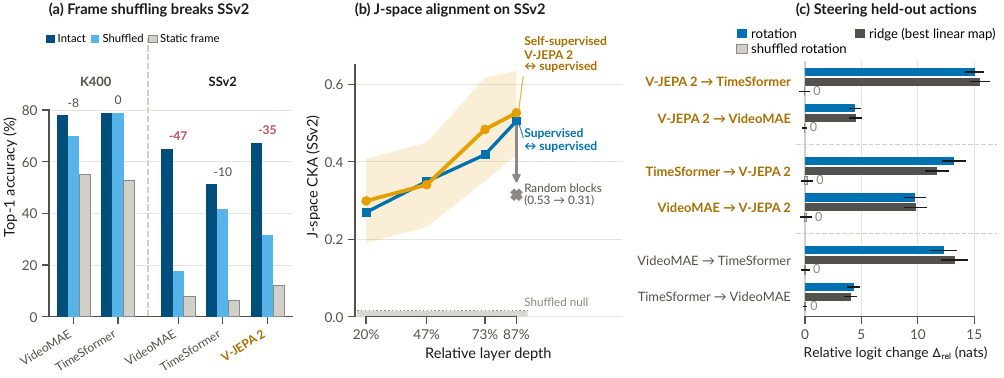}
\caption{\textbf{On SSv2, where frame order defines the action, self-supervised V-JEPA~2 aligns with supervised models in J-space and steers held-out actions through one rotation.} (a)~Top-1 accuracy on intact, frame-shuffled (change in points above the bars), and static-frame clips. (b)~SSv2 J-space debiased CKA across depth (mean and pair range) with random-block control and shuffled null. (c)~Held-out steering of 35 actions, measured as the target action's relative logit change $\Delta_{\mathrm{rel}}$ (main \cref{eq:rel}; 95\% intervals), for each of the six directed SSv2 pairs: rotation, ridge (best linear map), and shuffled rotation (grey).}
\label{fig:videossv2}
\end{figure*}

Kinetics-400 actions are largely recognizable from static appearance: shuffling frame order barely affects K400 accuracy (\cref{fig:videossv2}a) and leaves late J-space nearly unchanged (within-model intact-vs-shuffled CKA $\ge 0.97$).
A positive result on K400 alone therefore leaves open whether J-space captures genuine temporal dynamics rather than static visual cues.

Something-Something v2 (SSv2) eliminates such static shortcuts: shuffling or freezing frames collapses classification accuracy across every model (\cref{fig:videossv2}a).
Even in this strictly temporal regime, end-to-end VideoMAE-B and TimeSformer-B align \emph{more} strongly in late J-space on SSv2 than on K400 and steer held-out physical actions bidirectionally via a single orthogonal rotation (\cref{fig:videossv2}b, c).

\subsection{Self-supervised V-JEPA~2 aligns and steers like the supervised models}
\label{app:video-vjepa}
Across depth, V-JEPA~2-L aligns with the supervised SSv2 models in J-space as strongly as they align with one another, and re-initializing its encoder blocks drops this alignment sharply (\cref{fig:videossv2}b).
One rotation fitted on 139 actions steers the 35 held-out actions both \emph{into} and \emph{out of} V-JEPA~2-L, on par with ridge, while a rotation fitted on shuffled action pairs stays at zero (\cref{fig:videossv2}c).
Unlike the supervised models, V-JEPA~2-L's encoder never sees a label: only V-JEPA~2's classifier, released by Meta with the encoder, is trained on SSv2 actions, and the J-lens pulls each action's row back through it into the encoder (\cref{fig:videoarch}a).

\setcounter{figure}{\value{savedfigure}}\setcounter{equation}{\value{savedequation}}
\renewcommand{\thefigure}{S\arabic{figure}}\renewcommand{\theHfigure}{S\arabic{figure}}
\renewcommand{\theequation}{S\arabic{equation}}\renewcommand{\theHequation}{S\arabic{equation}}
\renewcommand{\thesubsection}{\thesection.\arabic{subsection}}\renewcommand{\theHsubsection}{\theHsection.\arabic{subsection}}
\setcounter{subsection}{0}

\appendix



\section{Agreement between language-model families under mutual $k$-NN}
\label{app:lmalign}

Replacing debiased CKA with normalized mutual 10-NN reproduces the same cross-family alignment ordering across depth and model scale (\cref{fig:maindepth,fig:mainscale}).

\subsection{Normalized mutual $k$-NN and scaling regression}
\label{app:mknn}

Let $N_k(s; K)$ denote the $k$ words most similar to word $s$ under kernel $K$ (excluding $s$ itself, with $k = 10$).
Normalized mutual $k$-NN \citep{huh2024platonic} averages the shared neighbor fraction $m$ over $\mathcal C$ and rescales it above the expected overlap $\kappa$ of a random ranking:
\begin{equation}
  m(K, K') \;=\; \frac{1}{|\mathcal C|}\sum_{s\in\mathcal C}\frac{|N_k(s; K)\cap N_k(s; K')|}{k},
  \label{eq:mknn-raw}
\end{equation}
\begin{equation}
  A(K, K') \;=\; \frac{m - \kappa}{1 - \kappa}, \qquad \kappa \;=\; \frac{k}{|\mathcal C| - 1}.
  \label{eq:mknn}
\end{equation}
By construction, $A = 1$ when two models assign identical $k$-nearest neighborhoods to every word, and $A = 0$ in expectation when one kernel's word labels are randomly permuted \citep{groger2026revisiting}.

Because models within the same family (\eg Llama or Pythia) share training corpora, tokenizers, and training recipes, every comparison is restricted to cross-family pairs.
Model $i$'s summary alignment is $\bar A_i = \operatorname{mean}_{j:\,F(j)\neq F(i)} A(K^{(i)}, K^{(j)})$, where $F(i)$ denotes family.
To measure scaling with parameter count $N_i$, we regress each cross-family pair's alignment $A_{ij}$ on its mean log parameter count with a separate intercept $a_{F(i)F(j)}$ per unordered family pair:
\begin{equation}
  A_{ij} \;=\; a_{F(i)F(j)} + \gamma \cdot \tfrac{1}{2}\big(\log_{10} N_i + \log_{10} N_j\big) + \varepsilon_{ij},
  \label{eq:slope}
\end{equation}
where $\gamma$ measures the alignment gain per tenfold increase in geometric-mean parameter count (95\% confidence intervals via model resampling).

\subsection{Global and local alignment across depth and scale}
\label{app:lmalign-depth}
\begin{figure}[t]
\centering
\includegraphics[width=\columnwidth]{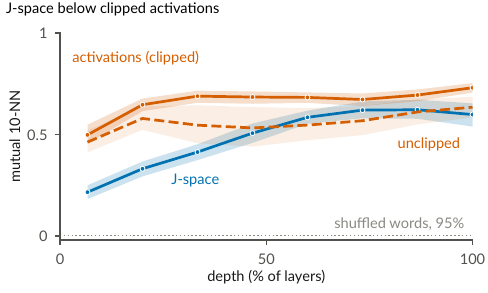}
\caption{\textbf{Under mutual 10-NN, cross-family J-space alignment rises with depth and matches unclipped activations.} Normalized mutual 10-NN (\cref{eq:mknn}, chance 0) by depth over 207 cross-family pairs of 23 language models, with 95\% model-bootstrap intervals: J-space (blue), clipped activations $K_h$ (solid orange), unclipped activations (dashed), and 95th-percentile shuffled null (dotted).}
\label{fig:maindepth}
\end{figure}

\begin{figure}[t]
\centering
\includegraphics[width=\columnwidth]{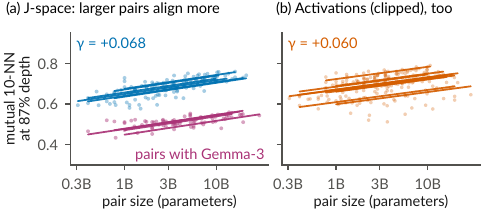}
\caption{\textbf{In late layers, larger language models align more under mutual 10-NN, in J-space and in activations.} Normalized mutual 10-NN at 87\% depth against each cross-family pair's geometric-mean parameter count in J-space (a; purple marks pairs involving Gemma-3) and clipped activations (b), with one line per family pair and shared slope $\gamma$ (\cref{eq:slope}, 95\% intervals $[0.042,0.101]$ and $[0.006,0.113]$).}
\label{fig:mainscale}
\end{figure}

J-space alignment rises with depth under both debiased CKA and mutual 10-NN. 
Under debiased CKA (\cref{fig:cka}a), cross-family J-space alignment rises from 0.50 at the first block to a 0.78 plateau across 60--87\% depth.
Mutual 10-NN (\cref{eq:mknn}) reproduces the same ordering across spaces (\cref{fig:maindepth}), and in late layers larger models align more closely: at 87\% depth, J-space alignment climbs by $\gamma = 0.068$ per tenfold parameter increase, mirroring clipped activations (\cref{fig:mainscale}) and rising within every family.
Under mutual 10-NN, J-space alignment also rises with model size.

\section{J-lens vectors steer their own model as J-space predicts}
\label{app:steer}

Before transferring a J-lens vector across models, we verify that each model's own averaged J-lens vector steers it as first-order theory predicts (\cref{eq:sidefx}; \cref{eq:avg-exact} in the \hyperref[app:prelim]{Preliminaries}).
Across 19 language models spanning all five families (410M--14B), adding $\alpha r_\ell \hat v_s$ at mid-depth (53\%, $\alpha = 1$, scoring 1,500 fixed candidates while steering 300 words) promotes $s$ to top-1 for 92--100\% of words in 18 of 19 models (Qwen3-8B climbs from 52\% at 53\% depth to 98.7\% at 73\% depth).
Across all 19 models, measured double-centered log-probability shifts on every other word (\cref{eq:dc}) correlate with the first-order J-space prediction $v_t^{\top}\hat v_s$ at mean Spearman 0.70, against 0.01--0.04 under shuffled words.

Steering every position $p \ge 16$ (\cref{eq:steer}) accumulates upstream pushes from all $p \le p'$, matching the J-lens average to first order (\cref{eq:avg-exact}), while every downstream normalization re-projects the perturbed residual stream onto a sphere of fixed norm, so a push tilts the direction that later blocks read rather than rescaling its length.


\section{Ridge regression and its rotational limit}
\label{app:ridge}

A pure rotation preserves lengths and angles, whereas an unconstrained linear map can stretch, shrink, and shear directions to repair distortion.
To measure how much of the transferable signal is purely rotational, we benchmark every Procrustes rotation (\cref{eq:rot}) against the best linear map fitted by ridge regression (\cref{eq:ridge}), referred to throughout as \emph{ridge}.
Both maps carry J-lens vectors from a source model $A$ to a \emph{receiver} $B$, the model that is then steered with the mapped vector.

To keep regularization comparable across widths and norms, we scale the penalty to the source variance,
\begin{equation}
  \lambda \;=\; \lambda_0\,\frac{\operatorname{tr}(G_A)}{|\mathcal F|}, \qquad G_A = \tilde X_A\tilde X_A^{\top},
  \label{eq:ridge-scale}
\end{equation}
choosing $\lambda_0$ by cross-validation to maximize held-out cosine with receiver $B$'s vectors (80/20 stem split over $\lambda_0 \in [10^{-3}, 10^{1}]$ between language models and up to $10^{3}$ into SDXL; 5-fold over fitting classes over $[10^{-2}, 10^{3}]$ for ViTs).

Heavy regularization pulls ridge directly toward our Procrustes rotation:
\begin{equation}
  \begin{aligned}
    W_{A\to B} &= \tilde Y_B^{\top}(G_A + \lambda I)^{-1}\tilde X_A \\
    &\xrightarrow{\;\lambda \to \infty\;} \frac{1}{\lambda}\,\tilde Y_B^{\top}\tilde X_A.
  \end{aligned}
  \label{eq:ridge-dual}
\end{equation}
Small $\lambda$ lets ridge invert $G_A$ to amplify low-variance source axes; large $\lambda$ suppresses that inversion, leaving $\tilde Y_B^{\top}\tilde X_A$---the exact cross-covariance whose orthogonal polar factor is our Procrustes rotation (\cref{eq:rot}).
In the strongly regularized ViT$\to$LM regime (\cref{app:vitsteer}), cross-validation selects large $\lambda_0$ near this limit (extending the grid to $10^{-4}$--$10^{8}$ and $\infty$ shifts gain by only 0.001 nats), where the orthogonal rotation retains 85\% of ridge's effect.


\section{Transfer by rotation: whitening and cross-model details}
\label{app:transfer}

For every receiver (the model steered with the mapped vector: a language model of another family, Qwen3-VL, or SDXL), we compare the J-space rotation with ridge (\cref{app:ridge}), the receiver's own J-lens vector, and two null controls: a rotation fitted on shuffled word pairs and a random direction.
Where a gap between plain rotation and ridge remains, unequal per-axis variance scales across models explain most of it.

\subsection{Per-model whitened rotation}
\label{app:transfer-white}

To separate per-model axis scaling from genuine cross-model shear, we shrink each model's empirical covariance $S_A = \tilde X_A^{\top}\tilde X_A/|\mathcal F|$ toward isotropic variance $\bar\sigma_A^2 = \operatorname{tr}(S_A)/d_A$,
\begin{equation}
  \begin{aligned}
    C_A &= (1-\eta)S_A + \eta\bar\sigma_A^2 I, \\
    W_{\mathrm{white}} &= C_B^{1/2}\,R\,C_A^{-1/2},
  \end{aligned}
  \label{eq:white}
\end{equation}
fitting the Procrustes rotation $R$ (\cref{eq:rot}) between whitened matrices $\tilde X_A C_A^{-1/2}$ and $\tilde Y_B C_B^{-1/2}$ (with shared shrinkage $\eta \in (0, 1]$ chosen on the same stem split as ridge; $\eta = 1$ recovers plain rotation).
This map uses only each model's own covariance and introduces no cross-model shear, yet closes most of the gap to ridge: it lifts LM$\to$LM side-effect Spearman from 0.72 to 0.77 at 87\% depth and 0.78 to 0.84 at 53\%, lifts Qwen3-VL at matched strength ($a = 0.05$) from 72.1\% to 81.9\% vs.\ 84.6\% for ridge, and lifts SDXL from 55\% to 60\%.
Removing the top 10 outlier directions instead lowers held-out cosine slightly ($-0.016$; \cref{app:why}, \cref{fig:why}b).

\subsection{Transfer between language models}
\label{app:transfer-lm}

We evaluate the 17 language models in four size tiers (56 directed cross-family pairs), scoring each mapped vector over 2,500 candidates (1,500 within-model words from \cref{app:steer} plus 1,000 stem-disjoint held-out words, the first 300 forming the test set) by its \emph{top-1 rate} and its \emph{side-effect Spearman} (Spearman correlation of double-centered shifts, \cref{eq:dc}, over all non-target candidates against the receiver's own vector).
Across the 56 late-layer (87\%) pairs, J-space rotation matches or \textit{beats} ridge on top-1 in 49 pairs (96.9--99.5\% across size tiers) and lifts the target's log-probability by 17.6 nats, equaling ridge (17.6) and the receiver's own vector (17.3).
At mid-depth (53\%), rotation leads ridge on top-1 by 95.5\% to 91.7\% (own vector: 98.2\% at 53\%, 95.2\% at 87\%).

A ridge map between raw unembedding rows ($J_\ell=I$), scored against the receiver's own unembedding row, reaches a side-effect Spearman of 0.68 at both 53\% and 87\% depth; in a depth sweep of leave-one-out ridge maps across 20 directed pairs of five 8--14B models, it drops to 0.44--0.48 at 20--33\% depth, whereas J-space leave-one-out ridge maps keep 0.85--0.86.

\subsection{Transfer to a multimodal receiver: Qwen3-VL}
\label{app:transfer-mm}

\begin{figure}[t]
\centering
\includegraphics[width=\linewidth]{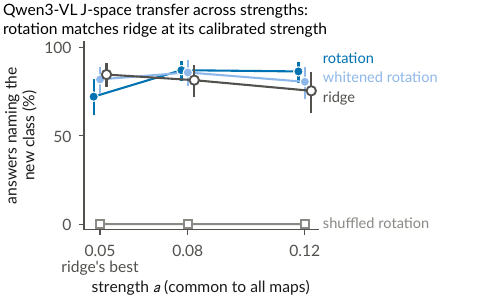}
\caption{\textbf{In Qwen3-VL, rotation's comparison with ridge depends on steering strength $a$, and per-model whitening closes the matched-$a$ gap.} Rate of naming the new class over 24 test pairs and four source LMs across shared strengths $a$ (\cref{eq:recv}, 95\% pair-bootstrap intervals), comparing J-space rotation, whitened rotation, ridge (peaking at $a=0.05$), and shuffled rotation.}
\label{fig:qwencontrols}
\end{figure}

\begin{figure}[t]
\centering
\includegraphics[width=\linewidth]{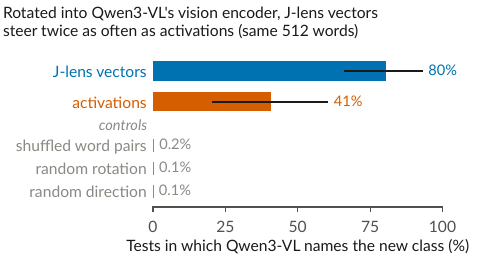}
\caption{\textbf{Rotated into Qwen3-VL's vision encoder, J-lens vectors steer about twice as often as activations.} For four source language models, a rotation fitted on the same 512 captioned words maps each word's J-lens vector onto Qwen3-VL's visual J-lens vectors (its answer readouts pulled back through $J_\ell$ into the vision encoder), or each word's mean last-token state onto Qwen3-VL's mean patch output on COCO images whose captions mention the word. Bars: share of 5,168 tests per map (24 pairs of 12 COCO classes) in which Qwen3-VL names the new class after the mapped vector is added to every patch; 95\% intervals resample classes. Gray: controls for the activation map (rotation fitted on shuffled word pairs, random rotation, random direction).}
\label{fig:qwenclean}
\end{figure}

\paragraph{Qwen3-VL strength tradeoff.}
In Qwen3-VL-2B (protocol in \cref{app:vision}), whether plain rotation or ridge names the new class more often depends on steering strength $a$ (\cref{fig:qwencontrols}).
Transferring from four language models (text-only Llama-3.1-8B, Pythia-12B, OLMo-2-13B, plus multimodal Gemma-3-12B), calibration on 15 pairs of 15 classes disjoint from the 12 test classes (smallest $a$ reaching 95\% of calibration peak) selects $a = 0.08$ for rotation and $a = 0.05$ for ridge ($a = 0.08$ for ridge from Gemma-3-12B).
At these calibrated strengths, J-space rotation names the new class on 87.1\% of tests (91.0\% across text-only sources; 88\% of Qwen3-VL's own vector at 99.0\%) against 83.9\% for ridge (main-text CIs resample the 12 test classes; bars in \cref{fig:qwencontrols,fig:qwenclean} resample pairs).

\paragraph{J-lens vectors against patch activations.}
On whole COCO images, a rotation fitted on 512 captioned words names the new class in 80.4\% of tests ($[66.0,93.1]$) from J-lens vectors, against 40.9\% ($[20.5,60.5]$) from activations (each word's mean last-token state in the source language model, and Qwen3-VL's mean final-vision-block patch output over COCO images whose captions mention it) (\cref{fig:qwenclean}).
Most of this gap comes from cluttered image targets: refitting every map on 155 concrete Open Images classes with activation targets from 32 clean single-object crops, J-space rotation reaches 58\%, clean-crop activation rotation 47\%, and cluttered-image activation rotation 33\%, with all controls $\le 0.9\%$.

\subsection{Transfer between language models and ViTs}
\label{app:transfer-vit}

Between language models and text-free ImageNet ViTs (\cref{app:vit,app:vitsteer}), rotation recovers 0.23 of the receiver's own J-lens vector's relative logit change $\Delta_{\mathrm{rel}}$ (\cref{eq:rel}) from language model to ViT (own vector 13.5 nats) and 0.20 in reverse (15.7 nats), against 0.27 and 0.23 for ridge (3.68 and 3.66 nats). Rotated activations recover 0.26 and 0.44, each relative to the receiver's own activation vector (9.64 and 6.89 nats); the ViT activations are mean patch-token activations.


\section{Geometry of the rotation: normalization, ReLU, and matched steering responses}
\label{app:why}

If the residual stream could be stretched or sheared at an intermediate layer without changing the model's output, angles between J-lens vectors at that layer would be arbitrary.
To see why, suppose an invertible $M \in \mathrm{GL}(d)$ could be applied to the residual stream, $h_\ell \mapsto M h_\ell$, with $M^{-1}$ absorbed into the weights that read it, so that every output is unchanged.
The J-lens vectors would then become $v_t \mapsto M^{-\top}v_t$ (main \cref{eq:basis}), and their inner products would become
\begin{equation}
  v_s^{\top}v_t \;\mapsto\; v_s^{\top}\big(M^{\top}M\big)^{-1}v_t .
  \label{eq:gauge-angles}
\end{equation}
Every angle survives only if $M^{-\top}$ acts on J-lens vectors as a rotation up to scale.
A stretch or shear would change the angles but not the model's function, so comparing two models' J-spaces would compare arbitrary choices of basis.

\begin{figure*}[t]
\centering
\includegraphics[width=\linewidth]{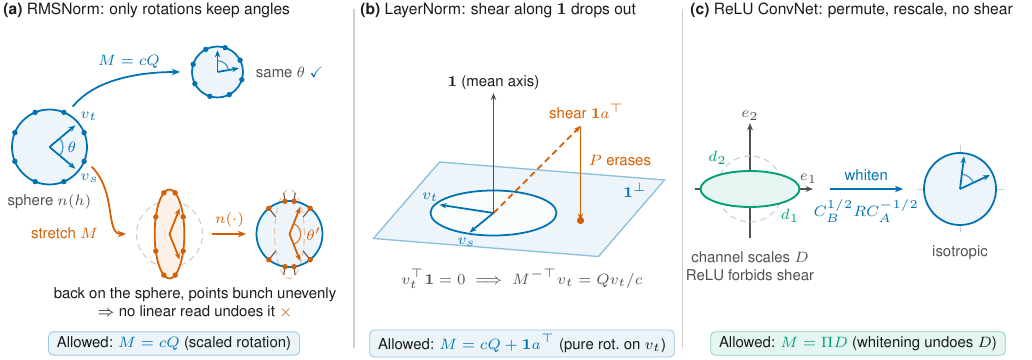}
\caption{\textbf{Each architecture's nonlinearity admits only a small group of basis changes, ruling out arbitrary cross-axis shear.} (a)~RMSNorm: a stretch $M$ turns the sphere into an ellipse, which $M^{-1}$ alone would undo; but renormalization rescales each point back onto the sphere by its own factor $\sqrt{d}/\|Mh\|$, so points bunch unevenly and no linear read undoes the result (a linear map of the sphere onto itself is a rotation, which moves every point evenly). Only scaled rotations $M=cQ$ survive. (b)~LayerNorm first projects onto the zero-mean hyperplane $\mathbf{1}^{\perp}$, erasing rank-one shears $\mathbf{1}a^{\top}$; because every J-lens vector satisfies $v_t^{\top}\mathbf{1}=0$, the action on J-lens vectors reduces to a pure scaled rotation $Qv_t/c$. (c)~ReLU ConvNet: a basis change commutes with the coordinate-wise ReLU only if it permutes and rescales channels, $M=\Pi D$, with no shear; per-model whitening ($C_B^{1/2}RC_A^{-1/2}$) makes the unidentified scales $D$ isotropic.}
\label{fig:normsym}
\end{figure*}

Yet J-space alignment rises with depth and model size and transfers across architectures and modalities.
The first step in explaining this is that the nonlinearities in the models we study admit almost no such $M$.
RMSNorm admits only a scaled rotation, and LayerNorm adds a shift that J-lens vectors do not see.
In a 3D ConvNet, the ReLU admits only channel permutations and per-channel scales, which per-model whitening removes (\cref{fig:normsym}).
We prove each below.
These constraints fix each model's angles but act on one model at a time, so they cannot by themselves place two models a rotation apart (main \cref{sec:why}).
The second half of this section shows that two models are a rotation apart exactly when their steering responses agree.

\paragraph{RMSNorm admits only scaled rotations.}
RMSNorm projects every state onto a sphere of radius $\sqrt{d}$, $n(h) = \sqrt{d}\,h/\|h\|$ (learned gain folded into the next linear read).
A stretch $M$ alone maps the sphere to an ellipsoid, which the linear read $M^{-1}$ would undo.
RMSNorm, however, rescales each point $Mh$ back onto the sphere by its own factor $\sqrt{d}/\|Mh\|$: directions are kept, but points of different length are rescaled by different amounts, so $n(Mh)$ is a nonlinear function of $n(h)$ that no linear read $T$ can reproduce (\cref{fig:normsym}a).
Formally, for some $c > 0$:
\begin{equation}
  n(Mh) = T\,n(h)\;\forall h \neq 0 \;\Longleftrightarrow\; M^{\top}\!M = c^2 I.
  \label{eq:orth}
\end{equation}
\emph{Proof.}
Expanding $n(Mh) = T\,n(h)$ gives $M^{-1}T\,h = (\|h\|/\|Mh\|)\,h$ for all $h \neq 0$.
Because a linear map sending every vector to a scalar multiple of itself must share one scalar across linearly independent vectors, $M^{-1}T = \mu I$ ($\mu > 0$), so $\|Mh\| = \|h\|/\mu$ everywhere.
Thus $h^{\top}(M^{\top}\!M - \mu^{-2}I)h = 0$ identically, forcing the symmetric matrix $M^{\top}\!M = c^2 I$ ($c = \mu^{-1}$).
Conversely, $M = cQ$ ($Q^{\top}Q = I$) gives $n(Mh) = Q\,n(h)$ \citep{ashkboos2024slicegpt}. $\square$
Because skip connections couple every layer to one shared residual basis (and $\epsilon > 0$ pins $c = 1$), even a single normalized read suffices.

\paragraph{LayerNorm adds a shear along $\mathbf{1}$ that drops out on J-lens vectors.}
LayerNorm first projects onto the zero-mean hyperplane $\mathbf{1}^{\perp}$ via $P = I - \tfrac{1}{d}\mathbf{1}\mathbf{1}^{\top}$, $n_{\mathrm{LN}}(h) = \sqrt{d}\,Ph/\|Ph\|$ (\cref{fig:normsym}b).
Requiring $n_{\mathrm{LN}}(Mh) = T\,n_{\mathrm{LN}}(h)$ forces $PM\mathbf{1} = 0$ (otherwise $n_{\mathrm{LN}}(M(\mathbf{1}+\varepsilon u))$ loses dependence on $u \in \mathbf{1}^{\perp}$ as $\varepsilon \to 0$), and the same quadratic-form argument on $\mathbf{1}^{\perp}$ yields all admissible maps:
\begin{equation}
  M \;=\; cQ + \mathbf{1}a^{\top} \qquad (c>0,\; Q^{\top}Q = I,\; Q\mathbf{1} = \mathbf{1}).
  \label{eq:ln-sym}
\end{equation}
Because every block reads through $P$ and folded readouts satisfy $u_t^{\top}\mathbf{1} = 0$ (\cref{eq:fold}), every J-lens vector lies flat inside $\mathbf{1}^{\perp}$ ($v_t^{\top}\mathbf{1} = 0$).
For any $v \in \mathbf{1}^{\perp}$, $M^{\top}(Qv) = (cQ^{\top} + a\mathbf{1}^{\top})Qv = cv + a(\mathbf{1}^{\top}v) = cv$, so $M^{-\top}v = Qv/c$---a pure scaled rotation with zero shear.

\paragraph{ReLU after each residual addition forbids cross-channel shear.}
The 3D ConvNet R(2+1)D-18 (\cref{app:video-method}) has no LayerNorm or RMSNorm, and at inference each \texttt{BatchNorm3d} is a fixed per-channel affine map that folds into the preceding convolution, so it constrains no basis change.
The constraint comes from the coordinate-wise ReLU $\sigma$ applied after each residual addition (torchvision's \texttt{BasicBlock} computes $h \mapsto \sigma(h + F(h))$): the state passed between blocks is a ReLU output, so a basis change $M \in \mathrm{GL}(C)$ of its $C$ channels can be absorbed into the adjacent convolutions only if it commutes with $\sigma$, and
\begin{equation}
  \sigma(Mx) = M\sigma(x)\;\forall x \;\Longleftrightarrow\; M = \Pi D,
  \label{eq:relu-sym}
\end{equation}
with $\Pi$ a channel permutation and $D$ positive diagonal \citep{godfrey2022symmetries}.
\emph{Proof.}
$\Pi D$ commutes with $\sigma$ because positive scaling preserves signs.
Conversely, $x = e_j$ gives $\sigma(m_j) = m_j$, so every column $m_j$ of $M$ is nonnegative, and $x = e_i - e_j$ gives $\sigma(m_i - m_j) = m_i$, so $m_j$ vanishes wherever $m_i > 0$.
The $C$ nonzero columns thus have disjoint supports, so each has exactly one positive entry, in a different row. $\square$
Cross-channel shear is therefore excluded, but the per-channel scales $D$ are not identified: each model's J-space carries a free diagonal gauge, which per-model whitening (\cref{eq:white}) removes without cross-model shear (\cref{fig:normsym}c).

\paragraph{Matching steering responses imply a rotation.}
A set of vectors is fixed up to a rotation by its pairwise inner products (its Gram matrix).
If two models' centered steering profiles $\tilde v_t^{\top}(\tilde v_s/\|\tilde v_s\|)$ (\cref{eq:sidefx}) match and their centered norms $\|\tilde v_s\|$ are proportional, their Gram matrices are proportional, $\tilde Y_B\tilde Y_B^{\top}=c^2\tilde X_A\tilde X_A^{\top}$, and conversely.
Two matrices with proportional Gram matrices share left singular vectors, with singular values in the ratio $c$, so matching their right singular vectors $V_A$ and $V_B$ gives the Procrustes rotation $R=V_BV_A^{\top}$ (\cref{eq:rot}) with $\tilde Y_B = c\,\tilde X_A R^{\top}$: a partial isometry that maps $\operatorname{span}(V_A)$ isometrically onto $\operatorname{span}(V_B)$. It may include reflections, and its rank is at most $\min(d_A, d_B, |\mathcal{F}|-1)$, because centering removes one dimension from the $|\mathcal{F}|$ fitting words; in the main paper's notation $R = UV^{\top}$ (\cref{eq:rot}), this matched subspace is $\operatorname{span}(V)$.

\begin{figure}[t]
\centering
\includegraphics[width=\linewidth]{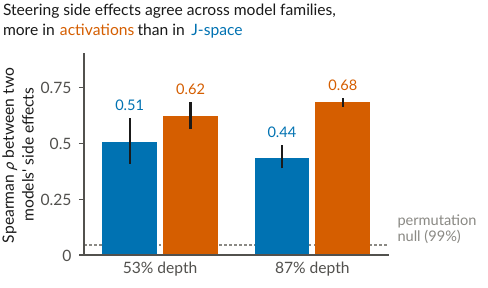}
\caption{\textbf{Steering side effects agree across language-model families, more in activations than in J-space.} Each of 17 language models steers 300 words with its own vector for each word (its J-lens vector, or its centered mean hidden state for activations), and the double-centered change in log-probability of about 2,200 other candidate words gives its side effects. Bars: Spearman correlation between two models' side effects, averaged over 115 pairs of models from different families (95\% intervals resample models). Dotted: largest per-pair 99th percentile when one model's word labels are shuffled.}
\label{fig:gram}
\end{figure}

That condition holds only approximately: steered by their own J-lens vectors, language models from different families shift other words alike at Spearman 0.44--0.51 in J-space, against 0.62--0.68 in activations (\cref{fig:gram}), all far above shuffled labels.

\begin{figure*}[t]
\centering
\includegraphics[width=\linewidth]{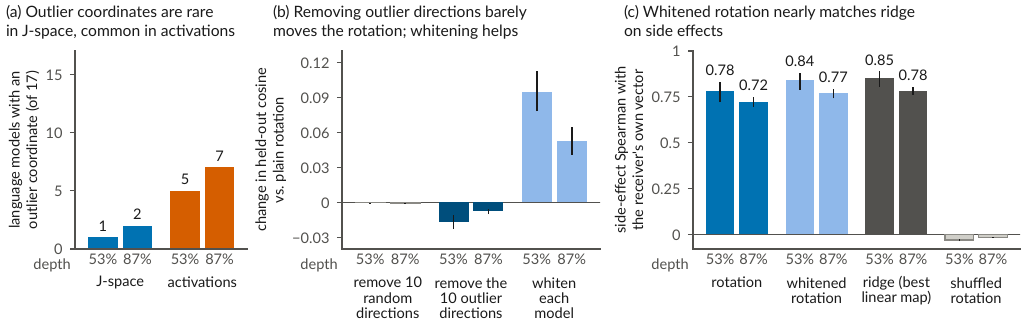}
\caption{\textbf{Whitening each model by its own covariance brings the rotation close to ridge; removing outlier directions does not.} 17 language models, 56 directed cross-family pairs, J-lens vectors at 53\% and 87\% depth. (a)~Models with an outlier coordinate (variance above $10\times$ the median and an eigenvector entry $\ge 0.5$): 1 and 2 of 17 in J-space (all Gemma-3), 5 and 7 in activations. (b)~Change in the rotation's held-out cosine with the receiver's own J-lens vector (300 held-out words) after removing 10 random directions per model, removing each model's top 10 directions, or whitening each model (\cref{eq:white}). (c)~Side-effect Spearman of the rotation, the whitened rotation, ridge (the best linear map) and a rotation fitted on shuffled word pairs. Error bars: 95\% bootstrap intervals over the 17 receivers.}
\label{fig:why}
\end{figure*}

\paragraph{Outlier coordinates and whitening.}
Ridge's lead over the rotation could come from a few outlier coordinates or from unequal variance along each model's own axes (\cref{fig:why}a--c).
Outlier coordinates (a coordinate variance above ten times the median and an eigenvector entry of at least 0.5) appear in J-space only in Gemma-3 (two of 17 models in late layers), but in the late-layer activations of 7 of 17 models (post hoc, on the 1,456 words of $\mathcal C$).
Removing the top 10 directions changes the rotation's held-out cosine with the receiver's own vector by $-0.016$ ($[-0.023,-0.011]$) at 53\% depth, whereas per-model whitening (\cref{eq:white}) raises it by $+0.095$ ($[0.078,0.113]$).

\begin{figure}[t]
\centering
\includegraphics[width=\linewidth]{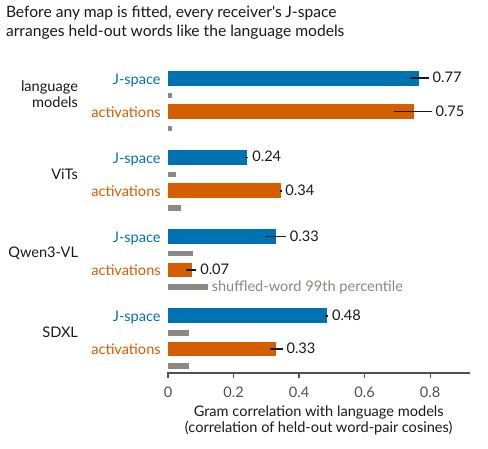}
\caption{\textbf{Before any map is fitted, every receiver's J-space arranges held-out words like the language models.} Gram correlation: the Pearson correlation, over pairs of held-out words, between their cosine similarities within the receiver and within a language model (0: unrelated; 1: identical). Gray: mean 99th percentile over 1,000 shuffles of the receiver's word labels; only Qwen3-VL's patch activations fall below it. Language models: 56 cross-family pairs of 17 models at 87\% depth, 300 held-out words; ViTs: 19 ViTs $\times$ 23 language models, 120 held-out classes; Qwen3-VL: 4 source models, 27 held-out classes; SDXL: 4 source models $\times$ 5 text-encoder blocks, 64 held-out queries. Error bars: 95\% bootstrap intervals.}
\label{fig:geomrecv}
\end{figure}

\paragraph{Every receiver arranges held-out words like the language models.}
Before any map is fitted, the cosine similarities among held-out words correlate across models above their 99th-percentile shuffled null for every receiver (0.24--0.77; 0.76 between language models on the fitting words; \cref{fig:geomrecv}); Qwen3-VL's J-space reaches 0.33, whereas its patch activations stay below the null (0.07 vs.\ 0.12).
A rotation then fits well: the Procrustes similarity $s=\|\tilde X_A^{\top}\tilde Y_B\|_*$ of the centered, unit-norm fitting matrices, which is 1 for an exact rotation \citep{williams2021generalized,ding2021grounding}, is 0.77 between language models in late layers against 0.40 for shuffled word pairs.

\section{Intermediate layers add agreement beyond the unembedding matrix}
\label{sec:tables}

Static unembedding matrices already agree across families: the cosine kernel $K_U$ of the effective readout vectors $u_t$ (\cref{eq:fold}) reaches 0.64 cross-family debiased CKA.
Regressing each model's $K_U$ out of its $K_J$ leaves a residual whose cross-family debiased CKA climbs from 0.40 in early blocks to 0.54 in late layers (vs.\ 0.002 shuffled), so the intermediate layers add agreement of their own.
Corpus statistics shared across training sets do not explain J-space agreement either: regressing co-occurrence and GloVe kernels out of each model's $K_J$ leaves late-layer debiased CKA at 0.72 (\cref{app:decomp}).


\section{Controls for skip-connection leakage and corpus statistics}
\label{app:decomp}

Neither skip-connection leakage nor corpus statistics shared across training sets accounts for cross-family J-space alignment.

\paragraph{Skip-connection leakage.}
Residual skip connections pass each effective readout vector $u_t$ (\cref{eq:fold}) through untrained blocks, so with random blocks $J_\ell \approx I$ and $v_t \approx u_t$: random-block alignment measures the trained output layer, not chance.
Re-initializing every transformer block to random weights while keeping the trained $W_E$, $W_U$, and final normalization (ten models) gives late-layer $K_J$ debiased CKA 0.57, against 0.78 with trained blocks ($+0.21$).
The part of $K_J$ that the unembedding does not explain also aligns across families (\cref{sec:tables}).
Over real hidden states, the basis-invariant kernel $K_S$ (\cref{app:states}) falls from 0.91 to 0.54 debiased CKA when random-block Jacobians replace the trained ones.

\paragraph{Corpus statistics.}
Regressing kernels built from positive pointwise mutual information over 100M WikiText-103 tokens and from 300-dimensional GloVe vectors \citep{pennington2014glove} out of each model's off-diagonal entries leaves late-layer J-space debiased CKA at 0.72 (0.78 raw).


\section{Qwen3-VL protocol, scaling, and visual perception probes}
\label{app:vision}

\paragraph{Visual J-lens and evaluation protocol.}
Because Qwen3-VL's vision encoder has no unembedding matrix into a fixed and discrete vocabulary, we differentiate the decoder's final pre-RMSNorm state (at the answer-predicting prompt position) back through the patch merger into the final vision block's 576 patches.
Qwen3-VL also passes features from intermediate vision-encoder layers directly into its first decoder layers; we hold those fixed.
Averaging this Jacobian over all patches, two one-word naming prompts, and COCO val2017 images (384$\times$384, excluding the 12 test classes) gives 1,446 stem-disjoint fitting vectors paired with 53\%-depth language-model vectors.
We evaluate 24 test pairs across 12 single-object COCO classes (30 photos/pair; \cref{fig:teaser}d and \cref{fig:demo}a show the first dog$\to$elephant photo and composite), calibrating strength $a$ (\cref{eq:recv}) on 15 pairs of 15 further classes, none of them among the 12 test classes (\cref{app:transfer-mm,fig:qwencontrols}).

In Qwen3-VL-2B, rotated 300-dimensional GloVe vectors \citep{pennington2014glove}, static word vectors learned from co-occurrence counts, name the new class on 50.8\% of tests, against 87.1\% for full-rank and 74.1\% for rank-300 J-space rotations.
Across Qwen3-VL-2B, 4B and 8B, J-space ridge maps (rotations were run only at 2B, where they reach 87.1\%, \cref{app:transfer}) name it on 84\%, 77\% and 77\%, about 6 points above ridge maps between static unembeddings ($J_\ell = I$: 78\%, 71\% and 71\%).

\begin{figure*}[t]
\centering
\includegraphics[width=\linewidth]{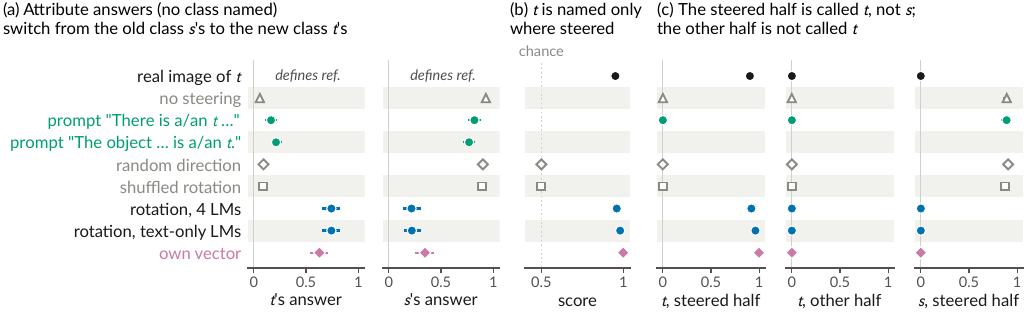}
\caption{\textbf{The rotated vector changes what Qwen3-VL reports about the object, but only in the half where it is added.} J-space rotations from four language models (LMs) over 24 pairs ($s\to t$, 95\% pair intervals). (a) Attribute answers matching Qwen3-VL's modal answer on a real image of $t$ (left) or $s$ (right); ``real image of $t$'' defines that reference (``\emph{defines ref.}''). (b) Half-steered score: 1 if $t$ is named for the steered half only, 0 for the other half only, 0.5 otherwise (dotted: chance). (c) Naming rates for $t$ (steered vs.\ unsteered half) and $s$ (steered half).}
\label{fig:vqa}
\end{figure*}

\paragraph{Probing visual perception (\cref{fig:vqa}).}
When we rotate steering vectors from text-only language models into Qwen3-VL, Qwen3-VL `sees' the new object (we steer towards) only in the region of the image where the steering vectors are added.
For example, on side-by-side composites of two $192\times 192$ photos of $s$ on a $384\times 384$ canvas, steering only the 12 patch columns of one half makes Qwen3-VL-2B name $t$ in the steered half alone on 91.8\% of composites, against 90.2\% for a real photo of $t$ in that half, with 0.1\% leakage into the unsteered half (\cref{fig:vqa}b, c).

If we ask 10 questions about physical attributes that do not name the class (\eg whether the object is alive, has legs or wheels, or can fly), the steered model answers as it does for a real image of $t$ on 74\% of the 95 questions whose answers differ between $s$ and $t$, against 6\% unsteered, 17--21\% with the word $t$ written in the prompt, and 9\% for shuffled or random vectors (\cref{fig:vqa}a).


\section{Contrastive image encoders agree in J-space only through their text towers}
\label{sec:vision}

In contrastive image encoders (four OpenCLIP ViTs trained on LAION-2B \citep{cherti2023reproducible,schuhmann2022laion} and three SigLIP2 towers \citep{tschannen2025siglip}, with $u_t$ the text tower's embedding of word $t$; 500 COCO \citep{lin2014microsoft} images, 296 object words), J-space alignment comes from the trained text tower rather than the image tower: random-weight image towers read through a trained text tower agree with one another (0.40--0.77) as much as trained ones (0.41--0.64).
With the text-tower and pooling-head kernels regressed out of $K_J$, the image tower's J-space aligns with the 23 language models at only 0.07--0.13 mutual 10-NN for CLIP and 0.02--0.03 for SigLIP2, below the 0.15 at which ViT-B image-tower activations align with them.


\section{Steering SDXL: protocol and further classes}
\label{app:gensteer}

Unlike a language model or VLM decoder, SDXL conditions its denoiser on penultimate representations from two frozen CLIP text encoders (768 and 1,280 dims).
To replace the generated object without disturbing \emph{a photo of a}, we split conditioning into five blocks (word-token mean and end-of-text/padding mean for each encoder, plus the larger encoder's pooled embedding) and fit one map per block from 87\%-depth LM J-lens vectors on 6,980 stem-disjoint words (\emph{a photo of a} + word).

\begin{figure}[t]
\centering
\includegraphics[width=\linewidth]{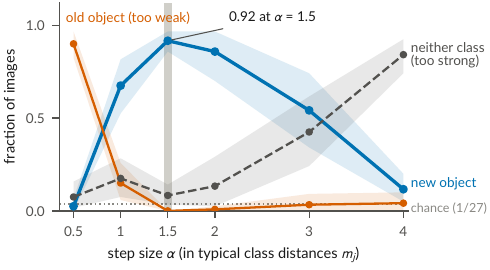}
\caption{\textbf{The SDXL step size $\alpha = 1.5$ (\cref{eq:sdxl-step}) maximizes the new-object rate on 15 calibration pairs that share no class with any test set.} Llama-3.1-8B ridge map at all five conditioning blocks; fraction of images whose nearest DINOv2 class mean (27 COCO classes, chance 1/27) is the new class, the old class (too weak), or neither (too strong); 8 seeds per pair (120 images per $\alpha$); bands are 95\% bootstrap intervals over pairs.}
\label{fig:sdxlalpha}
\end{figure}

\begin{figure*}[t]
\centering
\includegraphics[width=0.92\linewidth]{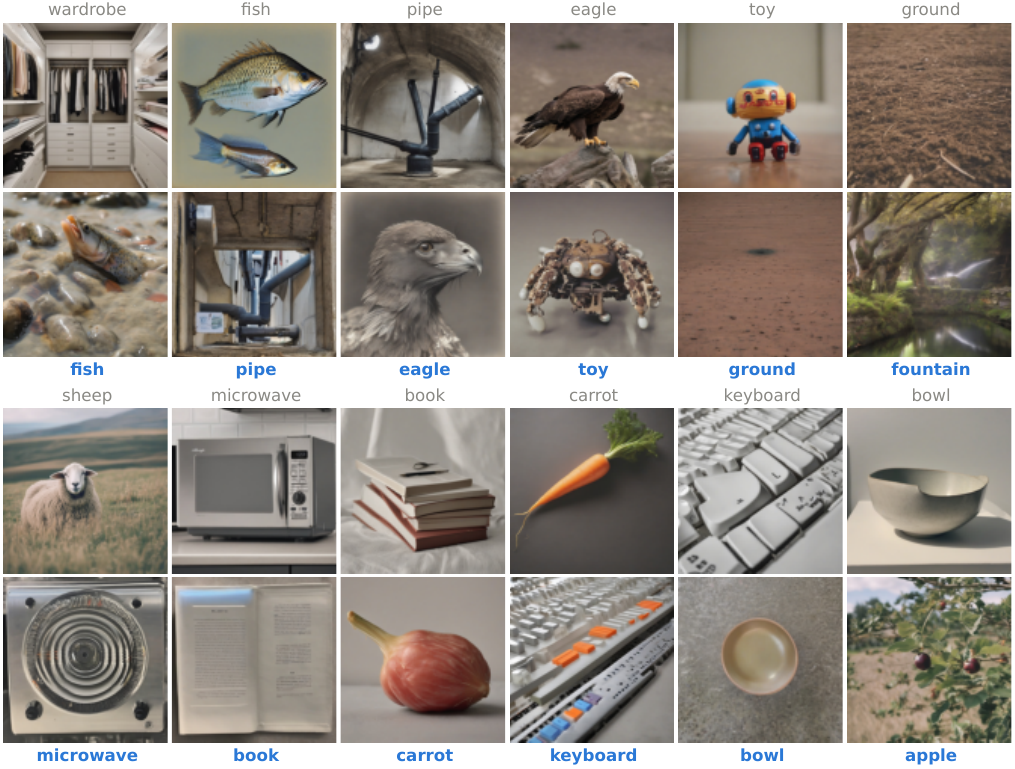}
\caption{\textbf{A J-lens vector from text-only Llama-3.1-8B, mapped by ridge, overrides the object SDXL generates.} Each column pairs an unsteered prompt (\emph{a photo of a} + gray word, top) with its same-seed steered output toward the blue word (bottom) for Open Images (upper rows) and COCO (lower rows): first 6 pairs per set at the first seed with Faster R-CNN person score $<0.05$.}
\label{fig:gensteer_supp}
\end{figure*}

Following \cref{eq:recv}, we add each mapped vector to the word, end-of-text/padding, and pooled positions while leaving the five prefix tokens untouched.
Because SDXL's five conditioning blocks differ in norm and spread, the step on block $j$ is measured in that block's typical distance between two classes:
\begin{equation}
  e_j \;\leftarrow\; e_j + \alpha\, m_j\, \frac{\delta_j}{\|\delta_j\|},
  \qquad m_j \;=\; \operatorname*{median}_{c \neq c'} \big\|e_j(c) - e_j(c')\big\|,
  \label{eq:sdxl-step}
\end{equation}
where $\delta_j$ is the mapped direction in block $j$, $e_j(c)$ is SDXL's own block-$j$ embedding of \emph{a photo of a} + $c$, and the median runs over all pairs of distinct calibration classes.
The strength $\alpha$ thus counts typical class distances: $\alpha = 1.5$ pushes the conditioning 1.5 class distances, and because the step uses the unit vector $\delta_j/\|\delta_j\|$, the same $\alpha$ gives the same step size whichever map produced the direction.
We set $\alpha$ once, at the peak of a sweep on 15 calibration pairs over 15 COCO classes that share no class with any test set, and hold it fixed for every test set and every map (\cref{fig:sdxlalpha}): too weak, and the old object survives; too strong, and the image shows neither object ($\alpha = 1.5$ and $2$ are within noise).
SDXL's own vector, the raw difference of its two class embeddings, is scaled by $\alpha = 2$, chosen by the same rule.
On the 40-class main set (60 pairs $\times$ 16 seeds, 30 steps, guidance 5, scored by nearest DINOv2 class mean), J-space rotations from four text-only LMs (Llama-3.1-8B, Pythia-12B, OLMo-2-7B, OLMo-2-32B) draw the new object in 51--57\% of images (mean 55\%), against 66\% for ridge, 60\% for whitened rotation and $\le 0.2\%$ for controls; \cref{fig:gensteer_supp} shows examples.
On all 229 remaining Open Images classes (343 pairs), refitted without test stems and with every hyperparameter held fixed, the rotations draw the new object in 26.3\% of images against 32.4\% for ridge (81\% of ridge).

\section{Text-free ImageNet ViTs align with language models}
\label{app:vit}

Supervised ImageNet Vision Transformers map pixels strictly to 1,000 integer class indices with zero text supervision (even though ImageNet's hierarchy originates from WordNet), yet pulling each ViT's classification head back through its Jacobian yields an internal J-space that aligns with all 23 language models.
We evaluate 19 supervised ViTs spanning DeiT, DeiT III, and AugReg (up to ViT-Huge, grouped into five recipe-by-pretraining families, including ImageNet-21k pretrained models) on the 161 single-token ImageNet class names shared by all 23 language tokenizers.
Adapting \cref{eq:jlens}, $h_L$ is the class token after the final LayerNorm and $u_c$ is the unmodified row $c$ of the classification head, so $u_c^{\top}J_\ell$ is the averaged gradient of logit $c$. The final LayerNorm thus sits inside the Jacobian instead of being folded into $u_c$ as in the \hyperref[app:prelim]{Preliminaries}; the two conventions agree to first order up to the per-image scale $1/\sigma$. Source positions are image patches averaged across 1,000 validation images (while $K_h$ averages the class token per class).
At 87\% depth, the 19 ViTs align with the 23 LMs at 0.245 mutual 10-NN and 0.336 debiased CKA in J-space (0.227 and 0.368 in activations), far above global (0.013) and within-category (0.068) shuffled nulls.


\section{Bidirectional steering between ImageNet ViTs and language models}
\label{app:vitsteer}

The J-space of a text-free ViT (\cref{app:vit}) also steers bidirectionally, including \emph{reverse steering} from ViTs directly into language models.
Across all 437 ViT--LM pairs, one class-agnostic rotation per direction (fitted on ImageNet classes excluding WordNet neighbors and near-duplicates of the 120 test classes) transfers positively in \emph{both} directions.
All depths were fixed before any steering result, with no reselection: ViT vectors are read and injected at 73\% depth, and 87\% depth, the language models' late depth used throughout the paper, was declared in advance as a second ViT depth to report; language-model vectors are read at 87\% depth and injected at 53\%, as in within-model steering (\cref{app:steer}).

\paragraph{Reverse steering (ViT $\to$ LM).}
J-lens vectors extracted from text-free ViTs directly steer language models: rotating 73\%-depth ViT vectors into 53\%-depth LM residual streams (at centered median residual norm across positions $p \ge 16$ of 16 held-out 128-token prompts) raises the target class name's relative logit $\Delta_{\mathrm{rel}}$ across all 437 pairs.
Net of shuffled controls ($-0.02$ to $0.00$), J-space rotation lifts the target by $+3.12$ nats and ridge by $+3.67$ nats uniformly across all 23 LMs (160M--32B) and all 19 ViTs, raising the target's mean prediction rank from 81 to 19.

\paragraph{Forward steering (LM $\to$ ViT) and controls.}
Conversely, injecting mapped 87\%-depth LM vectors into ViT patch tokens across 500 validation images lifts the target across all 437 pairs at both ViT depths: at 73\%, mean ridge lift is tightly clustered (3.40--3.98 nats) from 160M to 32B parameters; at 87\%, it is 3.87 nats (vs.\ 0.00 shuffled), pushing targets into top-5 on $>20\%$ of images and raising mean rank from 495 to 94.


\section{ViTs trained without normalization keep the rotation's share of ridge}
\label{app:dyt}

This section and \cref{app:persona,app:generation} test settings outside the main experiments: ViTs without normalization layers, persona vectors in instruction-tuned models, and generated multi-token text.
We train ViT-S/16 on ImageNet-1k twice with LayerNorm and twice with every LayerNorm replaced by the elementwise Dynamic Tanh (DyT) \citep{zhu2025transformers}, and steer as in \cref{app:vitsteer} across all 23 language models and 120 test classes.
In J-space, DyT minus LayerNorm on the rotation-to-ridge ratio of shuffle-subtracted $\Delta_{\mathrm{rel}}$ is $+0.10$ ($[0.06,0.13]$) from language model to ViT and $+0.03$ ($[-0.01,0.07]$) in reverse.
DyT's linear symmetries are signed permutations up to scale, themselves orthogonal, so this test cannot separate \cref{eq:orth} from other mechanisms that favor a rotation.


\section{A rotation fitted on words carries behavioral persona vectors}
\label{app:persona}

A rotation fitted only on isolated words carries behavioral persona vectors, which were never part of its fit, between instruction-tuned models.
Following \citet{chen2025persona}, we extract seven trait vectors as mean hidden-state differences over replies in Llama-3.1-8B-Instruct, Gemma-3-12B-it, OLMo-2-13B-Instruct, and Qwen3-14B, and carry each to another model by the activation rotation fitted on 6,000 words (excluding trait words and their stems).
Across 12 directed cross-family pairs, the mapped vector moves every testable trait in the direction of the receiver's own vector at 0.57 ($[0.38,0.78]$) of its effect.


\section{Self-supervised DINOv2/v3 read through a linear classifier}
\label{app:dino}

We trained an ImageNet linear classifier on top of each of 14 self-supervised ViTs, keeping their weights fixed: DINOv2 \citep{oquab2024dinov2} (small to giant, with and without registers \citep{darcet2024vision}) and DINOv3 \citep{simeoni2025dinov3} (small to 7B), and compared them with the 23 language models over the 161 class names of \cref{app:vit} at 87\% depth.
This is the only setting in the paper where the classifier is a single linear layer, trained by us on ImageNet, on top of an encoder never trained on labels; V-JEPA~2's classifier, by contrast, is a small attention-based classifier trained by Meta (\cref{app:video-vjepa}).
Their class-token activations align with the language models at 0.215 mutual 10-NN, while their J-space aligns at 0.140 (shuffled null 0.014), below a probe refitted on the same ViTs with randomly re-initialized blocks (0.174), and declines with scale ($-0.022$ per tenfold, $[-0.046, -0.002]$).


\section{Text generated under a visual prompt does not agree more with ViTs}
\label{app:generation}

Asking a language model to imagine seeing an object, rather than to describe it, barely changes its alignment with ViTs.
Following \citet{wang2026truth}, four instruction-tuned models (Qwen3-14B, Llama-3.1-8B-Instruct, Gemma-3-12B-it, and OLMo-2-13B-Instruct) generate 128 tokens after ``Imagine what it would look like to see a/an \emph{name}'' or, as a control, ``Tell me about a/an \emph{name}'', for the 161 class names and 19 ViTs of \cref{app:vit}.
Averaging 87\%-depth activations over the generated tokens, the visual prompt changes mutual 10-NN alignment with the ViTs by $-0.012$ ($[-0.033,0.008]$) relative to the control.
J-lenses fitted on each name's own transcripts gain only 0.009 after the visual prompt ($[-0.002,0.018]$; 0.016 by debiased CKA, $[0.003,0.028]$).


\section{Alignment and model size}
\label{app:size}

In late layers, J-space alignment grows with model size within families, mostly below roughly 1B parameters.
Regressing late-layer (87\%) debiased CKA across the 207 cross-family pairs on the two models' mean $\log_{10}$ parameter count, with one intercept per family pair (\cref{eq:slope}), gives a J-space slope of 0.052 per tenfold increase ($[0.022,0.087]$; 95\% intervals from resampling the 23 models).
This pooled slope moves the size of both models in a pair. A per-model slope, regressing each model's mean late-layer mutual 10-NN with the other families on its own $\log_{10}$ size within its family, is positive in all five families (0.010--0.050); because it moves only one model, it is about half the pooled mutual 10-NN slope ($\gamma = 0.068$, \cref{app:lmalign}). Most of the growth under debiased CKA (\cref{fig:cka}b) occurs below roughly 1B parameters, and Gemma-3 sits below every other family at all sizes (0.66--0.71 at 87\% depth).


\section{Final normalization and why the averaged J-lens works}
\label{app:lens}

The derivations the main paper cites here, folding the final normalization into the effective readout vector $u_t$ (\cref{eq:fold}) and the J-lens as a first-order context average (\cref{eq:avg-exact}), are in the \hyperref[app:prelim]{Preliminaries}, placed first because every section uses them.


\section{J-space alignment in video models}
\label{app:context}

The video results are in the \hyperref[app:videooverview]{video section} before Sec.~A.


\section{A basis-invariant J-space kernel over real hidden states}
\label{app:states}

Even without relying on normalization to fix Euclidean coordinates (\cref{app:why}), the scalar pairing $v_s^{\top}h_\ell$ is strictly invariant to \emph{any} invertible basis change $h_\ell \mapsto M h_\ell$ ($M \in \mathrm{GL}(d)$), since $(M^{-\top}v_s)^{\top}(M h_\ell) = v_s^{\top}h_\ell$.
Letting $\Sigma_h = \mathrm{Cov}(h_\ell)$ be the hidden-state covariance (estimated over 114,688 OpenWebText positions \citep{gokaslan2019openwebtext}, disjoint from WikiText-103, with compared models reading different passages), the linearized-readout covariance is strictly basis-invariant:
\begin{equation}
  \begin{aligned}
    \mathrm{Cov}(v_s^{\top}h_\ell,\, v_t^{\top}h_\ell) &= v_s^{\top}\Sigma_h v_t \\
    &= \big(\Sigma_h^{1/2}v_s\big)^{\top}\big(\Sigma_h^{1/2}v_t\big).
  \end{aligned}
  \label{eq:ks}
\end{equation}
We define $K_S$ as the cosine similarity matrix between word-centered rows of $V\Sigma_h^{1/2}$, weighting J-lens directions by how strongly real text excites them.

Across all 207 cross-family pairs in late layers (87\% depth), $K_S$ reaches 0.91 debiased CKA and 0.66 mutual 10-NN (unchanged on identical passages or after removing the top eigenvector of $\Sigma_h$), and collapses sharply whenever either the trained lens or trained states are replaced by random blocks (substituting $J_\ell = I$ on trained states drops CKA to 0.58, and random-block Jacobians on the same states to 0.54; \cref{app:decomp}).


\section{Robustness}
\label{app:robust}

Cross-family alignment is stable across word subsets, neighborhood sizes $k$, lens corpora, and numerical precision (95th-percentile shuffled null $< 0.002$ across all LM kernels).
Of the 40,000 most frequent English words, 13,369 meet the eligibility criteria of \cref{sec:setup}, and our 1,456-word vocabulary $\mathcal C$ samples 150 of them per frequency decile (removing 37 fragments and 7 low-snippet words: 939 nouns, 322 verbs, 145 adjectives, 50 adverbs); random draws (250--1,250 words), five disjoint 728-word splits, and varying $k \in [5, 50]$ shift mutual 10-NN by at most 0.011.
Switching the lens corpus from WikiText-103 to OpenWebText shifts alignment by at most 0.02 (Pythia-410M), split-half lenses agree at Spearman 0.98--1.00 across 14 models, and bfloat16 matches float32 within CKA $\ge 0.98$ (0.9999994 on Llama-3.1-8B; $\le 0.004$ alignment shift across models).
Hubness does not drive agreement: CSLS \citep{conneau2018word} and mutual proximity \citep{schnitzer2012local} shift alignment by at most 0.007.

%% file: main.bbl
\begin{thebibliography}{116}
\providecommand{\natexlab}[1]{#1}
\providecommand{\url}[1]{\texttt{#1}}
\expandafter\ifx\csname urlstyle\endcsname\relax
  \providecommand{\doi}[1]{doi: #1}\else
  \providecommand{\doi}{doi: \begingroup \urlstyle{rm}\Url}\fi

\bibitem[Agarwal(2026)]{agarwal2026cross}
Ayushi Agarwal.
\newblock Cross-architecture steering transfer in language models: A systematic
  empirical study.
\newblock \emph{arXiv preprint arXiv:2608.05164}, 2026.

\bibitem[Arnab et~al.(2021)Arnab, Dehghani, Heigold, Sun, Lu{\v{c}}i{\'c}, and
  Schmid]{arnab2021vivit}
Anurag Arnab, Mostafa Dehghani, Georg Heigold, Chen Sun, Mario Lu{\v{c}}i{\'c},
  and Cordelia Schmid.
\newblock {ViViT}: A video vision transformer.
\newblock In \emph{Proceedings of the IEEE/CVF International Conference on
  Computer Vision (ICCV)}, pages 6836--6846, 2021.

\bibitem[Artetxe et~al.(2018)Artetxe, Labaka, and
  Agirre]{artetxe2018generalizing}
Mikel Artetxe, Gorka Labaka, and Eneko Agirre.
\newblock Generalizing and improving bilingual word embedding mappings with a
  multi-step framework of linear transformations.
\newblock In \emph{Proceedings of the AAAI Conference on Artificial
  Intelligence}, pages 5012--5019, 2018.

\bibitem[Ashkboos et~al.(2024)Ashkboos, Croci, Gennari~do Nascimento, Hoefler,
  and Hensman]{ashkboos2024slicegpt}
Saleh Ashkboos, Maximilian~L. Croci, Marcelo Gennari~do Nascimento, Torsten
  Hoefler, and James Hensman.
\newblock {SliceGPT}: Compress large language models by deleting rows and
  columns.
\newblock In \emph{International Conference on Learning Representations
  (ICLR)}, 2024.

\bibitem[Assran et~al.(2025)Assran, Bardes, Fan, Garrido, Howes, Komeili,
  Muckley, Rizvi, Roberts, Sinha, Zholus, Arnaud, Gejji, Martin, Hogan, Dugas,
  Bojanowski, Khalidov, Labatut, Massa, Szafraniec, Krishnakumar, Li, Ma,
  Chandar, Meier, LeCun, Rabbat, and Ballas]{assran2025vjepa2}
Mido Assran, Adrien Bardes, David Fan, Quentin Garrido, Russell Howes, Mojtaba
  Komeili, Matthew Muckley, Ammar Rizvi, Claire Roberts, Koustuv Sinha, Artem
  Zholus, Sergio Arnaud, Abha Gejji, Ada Martin, Francois~Robert Hogan, Daniel
  Dugas, Piotr Bojanowski, Vasil Khalidov, Patrick Labatut, Francisco Massa,
  Marc Szafraniec, Kapil Krishnakumar, Yong Li, Xiaodong Ma, Sarath Chandar,
  Franziska Meier, Yann LeCun, Michael Rabbat, and Nicolas Ballas.
\newblock {V-JEPA 2}: Self-supervised video models enable understanding,
  prediction and planning.
\newblock \emph{arXiv preprint arXiv:2506.09985}, 2025.

\bibitem[Baars(1988)]{Baars}
Bernard~J. Baars.
\newblock \emph{A Cognitive Theory of Consciousness}.
\newblock Cambridge University Press, 1988.

\bibitem[Bai et~al.(2025)Bai, Cai, Chen, et~al.]{bai2025qwen3vl}
Shuai Bai, Yuxuan Cai, Ruizhe Chen, et~al.
\newblock {Qwen3-VL} technical report.
\newblock \emph{arXiv preprint arXiv:2511.21631}, 2025.

\bibitem[Balestriero and LeCun(2026)]{balestriero2026proving}
Randall Balestriero and Yann LeCun.
\newblock Proving and improving the {Platonic} representation hypothesis with
  {JEPAs}.
\newblock \emph{arXiv preprint arXiv:2610.39302}, 2026.

\bibitem[Bansal et~al.(2021)Bansal, Nakkiran, and Barak]{bansal2021revisiting}
Yamini Bansal, Preetum Nakkiran, and Boaz Barak.
\newblock Revisiting model stitching to compare neural representations.
\newblock In \emph{Advances in Neural Information Processing Systems
  (NeurIPS)}, 2021.

\bibitem[Baumann et~al.(2025)Baumann, Krause, Neumayr, Stracke, Sevi, Hu, and
  Ommer]{baumann2025continuous}
Stefan~Andreas Baumann, Felix Krause, Michael Neumayr, Nick Stracke, Melvin
  Sevi, Vincent~Tao Hu, and Bj{\"o}rn Ommer.
\newblock Continuous, subject-specific attribute control in {T2I} models by
  identifying semantic directions.
\newblock In \emph{Proceedings of the IEEE/CVF Conference on Computer Vision
  and Pattern Recognition (CVPR)}, 2025.

\bibitem[Belrose et~al.(2023)Belrose, Ostrovsky, McKinney, Furman, Smith,
  Halawi, Biderman, and Steinhardt]{belrose2023eliciting}
Nora Belrose, Igor Ostrovsky, Lev McKinney, Zach Furman, Logan Smith, Danny
  Halawi, Stella Biderman, and Jacob Steinhardt.
\newblock Eliciting latent predictions from transformers with the tuned lens.
\newblock \emph{arXiv preprint arXiv:2303.08112}, 2023.

\bibitem[Bertasius et~al.(2021)Bertasius, Wang, and
  Torresani]{bertasius2021spacetime}
Gedas Bertasius, Heng Wang, and Lorenzo Torresani.
\newblock Is space-time attention all you need for video understanding?
\newblock In \emph{Proceedings of the 38th International Conference on Machine
  Learning (ICML)}, pages 813--824, 2021.

\bibitem[Biderman et~al.(2023)Biderman, Schoelkopf, Anthony,
  et~al.]{biderman2023pythia}
Stella Biderman, Hailey Schoelkopf, Quentin Anthony, et~al.
\newblock Pythia: A suite for analyzing large language models across training
  and scaling.
\newblock In \emph{Proceedings of the 40th International Conference on Machine
  Learning (ICML)}, pages 2397--2430, 2023.

\bibitem[Bosco and Srinivasan(2026)]{bosco2026locating}
Preethi~Carmel Bosco and Gopalakrishnan Srinivasan.
\newblock Locating and steering refusal beyond attention.
\newblock \emph{arXiv preprint arXiv:2609.04721}, 2026.

\bibitem[Cardenas~Piepereit(2026)]{cardenaspiepereit2026architecture}
Fernando Cardenas~Piepereit.
\newblock Architecture-dependent causal transfer of activation states across
  large language models.
\newblock \emph{arXiv preprint arXiv:2608.16347}, 2026.

\bibitem[Changpinyo et~al.(2016)Changpinyo, Chao, Gong, and
  Sha]{changpinyo2016synthesized}
Soravit Changpinyo, Wei-Lun Chao, Boqing Gong, and Fei Sha.
\newblock Synthesized classifiers for zero-shot learning.
\newblock In \emph{Proceedings of the IEEE Conference on Computer Vision and
  Pattern Recognition (CVPR)}, pages 5327--5336, 2016.

\bibitem[Chen et~al.(2025{\natexlab{a}})Chen, Merullo, Stolfo, and
  Pavlick]{chen2025transferring}
Alan Chen, Jack Merullo, Alessandro Stolfo, and Ellie Pavlick.
\newblock Transferring linear features across language models with model
  stitching.
\newblock In \emph{Advances in Neural Information Processing Systems
  (NeurIPS)}, 2025{\natexlab{a}}.

\bibitem[Chen et~al.(2025{\natexlab{b}})Chen, Arditi, Sleight, Evans, and
  Lindsey]{chen2025persona}
Runjin Chen, Andy Arditi, Henry Sleight, Owain Evans, and Jack Lindsey.
\newblock Persona vectors: Monitoring and controlling character traits in
  language models.
\newblock \emph{arXiv preprint arXiv:2507.21509}, 2025{\natexlab{b}}.

\bibitem[Cherti et~al.(2023)Cherti, Beaumont, Wightman,
  et~al.]{cherti2023reproducible}
Mehdi Cherti, Romain Beaumont, Ross Wightman, et~al.
\newblock Reproducible scaling laws for contrastive language-image learning.
\newblock In \emph{Proceedings of the IEEE/CVF Conference on Computer Vision
  and Pattern Recognition (CVPR)}, pages 2818--2829, 2023.

\bibitem[Conneau et~al.(2018)Conneau, Lample, Ranzato, Denoyer, and
  J{\'e}gou]{conneau2018word}
Alexis Conneau, Guillaume Lample, Marc'Aurelio Ranzato, Ludovic Denoyer, and
  Herv{\'e} J{\'e}gou.
\newblock Word translation without parallel data.
\newblock In \emph{International Conference on Learning Representations
  (ICLR)}, 2018.

\bibitem[Csiszárik et~al.(2021)Csiszárik, Kőrösi-Szabó, Matszangosz, Papp,
  and Varga]{csiszarik2021similarity}
Adrián Csiszárik, Péter Kőrösi-Szabó, {\'A}kos~K. Matszangosz, Gergely
  Papp, and Dániel Varga.
\newblock Similarity and matching of neural network representations.
\newblock In \emph{Advances in Neural Information Processing Systems
  (NeurIPS)}, 2021.

\bibitem[Darcet et~al.(2024)Darcet, Oquab, Mairal, and
  Bojanowski]{darcet2024vision}
Timoth{\'e}e Darcet, Maxime Oquab, Julien Mairal, and Piotr Bojanowski.
\newblock Vision transformers need registers.
\newblock In \emph{International Conference on Learning Representations
  (ICLR)}, 2024.

\bibitem[Dehaene and Naccache(2001)]{dehaene}
Stanislas Dehaene and Lionel Naccache.
\newblock Towards a cognitive neuroscience of consciousness: Basic evidence and
  a workspace framework.
\newblock \emph{Cognition}, 79\penalty0 (1--2):\penalty0 1--37, 2001.

\bibitem[Deng et~al.(2009)Deng, Dong, Socher, Li, Li, and
  Fei-Fei]{deng2009imagenet}
Jia Deng, Wei Dong, Richard Socher, Li-Jia Li, Kai Li, and Li Fei-Fei.
\newblock {ImageNet}: A large-scale hierarchical image database.
\newblock In \emph{Proceedings of the IEEE Conference on Computer Vision and
  Pattern Recognition (CVPR)}, pages 248--255, 2009.

\bibitem[Ding et~al.(2021)Ding, Denain, and Steinhardt]{ding2021grounding}
Frances Ding, Jean-Stanislas Denain, and Jacob Steinhardt.
\newblock Grounding representation similarity through statistical testing.
\newblock In \emph{Advances in Neural Information Processing Systems
  (NeurIPS)}, 2021.

\bibitem[Dosovitskiy et~al.(2021)Dosovitskiy, Beyer, Kolesnikov,
  et~al.]{dosovitskiy2021image}
Alexey Dosovitskiy, Lucas Beyer, Alexander Kolesnikov, et~al.
\newblock An image is worth 16x16 words: Transformers for image recognition at
  scale.
\newblock In \emph{International Conference on Learning Representations
  (ICLR)}, 2021.

\bibitem[Elhage et~al.(2021)Elhage, Nanda, Olsson, Henighan, Joseph, Mann,
  Askell, Bai, Chen, Conerly, DasSarma, Drain, Ganguli, Hatfield-Dodds,
  Hernandez, Jones, Kernion, Lovitt, Ndousse, Amodei, Brown, Clark, Kaplan,
  McCandlish, and Olah]{elhage2021mathematical}
Nelson Elhage, Neel Nanda, Catherine Olsson, Tom Henighan, Nicholas Joseph, Ben
  Mann, Amanda Askell, Yuntao Bai, Anna Chen, Tom Conerly, Nova DasSarma, Dawn
  Drain, Deep Ganguli, Zac Hatfield-Dodds, Danny Hernandez, Andy Jones, Jackson
  Kernion, Liane Lovitt, Kamal Ndousse, Dario Amodei, Tom Brown, Jack Clark,
  Jared Kaplan, Sam McCandlish, and Chris Olah.
\newblock A mathematical framework for transformer circuits.
\newblock Transformer Circuits Thread,
  \url{https://transformer-circuits.pub/2021/framework/index.html}, 2021.

\bibitem[Elhage et~al.(2023)Elhage, Lasenby, and Olah]{elhage2023privileged}
Nelson Elhage, Robert Lasenby, and Christopher Olah.
\newblock Privileged bases in the transformer residual stream.
\newblock Transformer Circuits Thread,
  \url{https://transformer-circuits.pub/2023/privileged-basis/}, 2023.

\bibitem[Frome et~al.(2013)Frome, Corrado, Shlens, Bengio, Dean, Ranzato, and
  Mikolov]{frome2013devise}
Andrea Frome, Greg~S. Corrado, Jonathon Shlens, Samy Bengio, Jeffrey Dean,
  Marc'Aurelio Ranzato, and Tomas Mikolov.
\newblock {DeViSE}: A deep visual-semantic embedding model.
\newblock In \emph{Advances in Neural Information Processing Systems
  (NeurIPS)}, pages 2121--2129, 2013.

\bibitem[Gan et~al.(2026{\natexlab{a}})Gan, Li, and Luo]{gan2026beyond}
Jin Gan, Xin Li, and Jun Luo.
\newblock Beyond token-level guidance: Inference-time alignment of specialized
  {LLMs} via cross-family representation steering.
\newblock In \emph{Proceedings of the Conference on Empirical Methods in
  Natural Language Processing (EMNLP)}, 2026{\natexlab{a}}.
\newblock arXiv:2608.30319.

\bibitem[Gan et~al.(2026{\natexlab{b}})Gan, Fu, Asilis, Liu, Yogatama, Sharan,
  Jia, and Neiswanger]{gan2026textual}
Woody~Haosheng Gan, Deqing Fu, Julian Asilis, Ollie Liu, Dani Yogatama, Vatsal
  Sharan, Robin Jia, and Willie Neiswanger.
\newblock Textual steering vectors can improve visual understanding in
  multimodal large language models.
\newblock In \emph{Proceedings of the 64th Annual Meeting of the Association
  for Computational Linguistics (Volume 1: Long Papers)}, 2026{\natexlab{b}}.

\bibitem[Gandelsman et~al.(2024)Gandelsman, Efros, and
  Steinhardt]{gandelsman2024interpreting}
Yossi Gandelsman, Alexei~A. Efros, and Jacob Steinhardt.
\newblock Interpreting {CLIP}'s image representation via text-based
  decomposition.
\newblock In \emph{International Conference on Learning Representations
  (ICLR)}, 2024.

\bibitem[Gandikota et~al.(2024)Gandikota, Materzynska, Zhou, Torralba, and
  Bau]{gandikota2024concept}
Rohit Gandikota, Joanna Materzynska, Tingrui Zhou, Antonio Torralba, and David
  Bau.
\newblock Concept sliders: {LoRA} adaptors for precise control in diffusion
  models.
\newblock In \emph{European Conference on Computer Vision (ECCV)}, 2024.

\bibitem[{Gemma Team} et~al.(2025){Gemma Team}, Kamath, Ferret,
  et~al.]{gemmateam2025gemma}
{Gemma Team}, Aishwarya Kamath, Johan Ferret, et~al.
\newblock {Gemma} 3 technical report.
\newblock \emph{arXiv preprint arXiv:2503.19786}, 2025.

\bibitem[Godfrey et~al.(2022)Godfrey, Brown, Emerson, and
  Kvinge]{godfrey2022symmetries}
Charles Godfrey, Davis Brown, Tegan Emerson, and Henry Kvinge.
\newblock On the symmetries of deep learning models and their internal
  representations.
\newblock In \emph{Advances in Neural Information Processing Systems
  (NeurIPS)}, pages 11893--11905, 2022.

\bibitem[Gokaslan and Cohen(2019)]{gokaslan2019openwebtext}
Aaron Gokaslan and Vanya Cohen.
\newblock {OpenWebText} corpus.
\newblock \url{http://Skylion007.github.io/OpenWebTextCorpus}, 2019.
\newblock Accessed 2026-10-31.

\bibitem[Goyal et~al.(2017)Goyal, Ebrahimi~Kahou, Michalski, Materzynska,
  Westphal, Kim, Haenel, Fruend, Yianilos, Mueller-Freitag, Hoppe, Thurau, Bax,
  and Memisevic]{goyal2017something}
Raghav Goyal, Samira Ebrahimi~Kahou, Vincent Michalski, Joanna Materzynska,
  Susanne Westphal, Heuna Kim, Valentin Haenel, Ingo Fruend, Peter Yianilos,
  Moritz Mueller-Freitag, Florian Hoppe, Christian Thurau, Ingo Bax, and Roland
  Memisevic.
\newblock The ``something something'' video database for learning and
  evaluating visual common sense.
\newblock In \emph{Proceedings of the IEEE International Conference on Computer
  Vision (ICCV)}, pages 5842--5850, 2017.

\bibitem[Grattafiori et~al.(2024)Grattafiori, Dubey, Jauhri,
  et~al.]{grattafiori2024llama}
Aaron Grattafiori, Abhimanyu Dubey, Abhinav Jauhri, et~al.
\newblock The {Llama} 3 herd of models.
\newblock \emph{arXiv preprint arXiv:2407.21783}, 2024.

\bibitem[Gröger et~al.(2026)Gröger, Wen, and Brbić]{groger2026revisiting}
Fabian Gröger, Shuo Wen, and Maria Brbić.
\newblock Revisiting the {Platonic} representation hypothesis: An
  {Aristotelian} view.
\newblock In \emph{Proceedings of the 43rd International Conference on Machine
  Learning (ICML)}, pages 36724--36766, 2026.

\bibitem[Gurnee et~al.(2026)Gurnee, Sofroniew, Pearce,
  et~al.]{gurnee2026verbalizable}
Wes Gurnee, Nicholas Sofroniew, Adam Pearce, et~al.
\newblock Verbalizable representations form a global workspace in language
  models.
\newblock \emph{arXiv preprint arXiv:2607.15495}, 2026.
\newblock Also published on the Transformer Circuits Thread, 6 Jul 2026.

\bibitem[Huang et~al.(2025)Huang, Huang, Feng, Lei, and Lv]{huang2025cross}
Youcheng Huang, Chen Huang, Duanyu Feng, Wenqiang Lei, and Jiancheng Lv.
\newblock Cross-model transferability among large language models on the
  {Platonic} representations of concepts.
\newblock In \emph{Proceedings of the 63rd Annual Meeting of the Association
  for Computational Linguistics (Volume 1: Long Papers)}, pages 3686--3704,
  2025.

\bibitem[Huh et~al.(2024)Huh, Cheung, Wang, and Isola]{huh2024platonic}
Minyoung Huh, Brian Cheung, Tongzhou Wang, and Phillip Isola.
\newblock Position: The {Platonic} representation hypothesis.
\newblock In \emph{Proceedings of the 41st International Conference on Machine
  Learning (ICML)}, pages 20617--20642, 2024.

\bibitem[Jha et~al.(2025)Jha, Zhang, Shmatikov, and Morris]{jha2025harnessing}
Rishi Jha, Collin Zhang, Vitaly Shmatikov, and John~X. Morris.
\newblock Harnessing the universal geometry of embeddings.
\newblock In \emph{Advances in Neural Information Processing Systems
  (NeurIPS)}, pages 45963--45987, 2025.

\bibitem[Jiang et~al.(2025)Jiang, Kachinthaya, Petryk, and
  Gandelsman]{jiang2025interpreting}
Nick Jiang, Anish Kachinthaya, Suzie Petryk, and Yossi Gandelsman.
\newblock Interpreting and editing vision-language representations to mitigate
  hallucinations.
\newblock In \emph{International Conference on Learning Representations
  (ICLR)}, 2025.

\bibitem[Kapoor et~al.(2025)Kapoor, Srivastava, and Khosla]{kapoor2025bridging}
Chaitanya Kapoor, Sudhanshu Srivastava, and Meenakshi Khosla.
\newblock Bridging critical gaps in convergent learning: How representational
  alignment evolves across layers, training, and distribution shifts.
\newblock In \emph{Advances in Neural Information Processing Systems
  (NeurIPS)}, 2025.

\bibitem[Kay et~al.(2017)Kay, Carreira, Simonyan, Zhang, Hillier,
  Vijayanarasimhan, Viola, Green, Back, Natsev, Suleyman, and
  Zisserman]{kay2017kinetics}
Will Kay, Jo{\~a}o Carreira, Karen Simonyan, Brian Zhang, Chloe Hillier,
  Sudheendra Vijayanarasimhan, Fabio Viola, Tim Green, Trevor Back, Paul
  Natsev, Mustafa Suleyman, and Andrew Zisserman.
\newblock The {Kinetics} human action video dataset.
\newblock \emph{arXiv preprint arXiv:1705.06950}, 2017.

\bibitem[Kornblith et~al.(2019)Kornblith, Norouzi, Lee, and
  Hinton]{kornblith2019similarity}
Simon Kornblith, Mohammad Norouzi, Honglak Lee, and Geoffrey Hinton.
\newblock Similarity of neural network representations revisited.
\newblock In \emph{Proceedings of the 36th International Conference on Machine
  Learning (ICML)}, pages 3519--3529, 2019.

\bibitem[Kovaleva et~al.(2021)Kovaleva, Kulshreshtha, Rogers, and
  Rumshisky]{kovaleva2021bert}
Olga Kovaleva, Saurabh Kulshreshtha, Anna Rogers, and Anna Rumshisky.
\newblock {BERT} busters: Outlier dimensions that disrupt transformers.
\newblock In \emph{Findings of the Association for Computational Linguistics:
  ACL-IJCNLP 2021}, 2021.

\bibitem[Kriegeskorte et~al.(2008)Kriegeskorte, Mur, and
  Bandettini]{kriegeskorte2008representational}
Nikolaus Kriegeskorte, Marieke Mur, and Peter Bandettini.
\newblock Representational similarity analysis -- connecting the branches of
  systems neuroscience.
\newblock \emph{Frontiers in Systems Neuroscience}, 2:\penalty0 4, 2008.

\bibitem[Kuznetsova et~al.(2020)Kuznetsova, Rom, Alldrin, Uijlings, Krasin,
  Pont-Tuset, Kamali, Popov, Malloci, Kolesnikov, Duerig, and
  Ferrari]{kuznetsova2020open}
Alina Kuznetsova, Hassan Rom, Neil Alldrin, Jasper Uijlings, Ivan Krasin, Jordi
  Pont-Tuset, Shahab Kamali, Stefan Popov, Matteo Malloci, Alexander
  Kolesnikov, Tom Duerig, and Vittorio Ferrari.
\newblock The {Open Images} dataset {V4}: Unified image classification, object
  detection, and visual relationship detection at scale.
\newblock \emph{International Journal of Computer Vision}, 128\penalty0
  (7):\penalty0 1956--1981, 2020.

\bibitem[L{\"a}hner and Moeller(2024)]{lahner2024direct}
Zorah L{\"a}hner and Michael Moeller.
\newblock On the direct alignment of latent spaces.
\newblock In \emph{Proceedings of UniReps: the First Workshop on Unifying
  Representations in Neural Models}, pages 158--169, 2024.

\bibitem[Lee et~al.(2025)Lee, Weber, Viégas, and Wattenberg]{lee2025shared}
Andrew Lee, Melanie Weber, Fernanda Viégas, and Martin Wattenberg.
\newblock Shared global and local geometry of language model embeddings.
\newblock In \emph{Conference on Language Modeling (COLM)}, 2025.

\bibitem[Lenc and Vedaldi(2015)]{lenc2015understanding}
Karel Lenc and Andrea Vedaldi.
\newblock Understanding image representations by measuring their equivariance
  and equivalence.
\newblock In \emph{Proceedings of the IEEE Conference on Computer Vision and
  Pattern Recognition (CVPR)}, pages 991--999, 2015.

\bibitem[Li et~al.(2016)Li, Yosinski, Clune, Lipson, and
  Hopcroft]{li2016convergent}
Yixuan Li, Jason Yosinski, Jeff Clune, Hod Lipson, and John Hopcroft.
\newblock Convergent learning: Do different neural networks learn the same
  representations?
\newblock In \emph{International Conference on Learning Representations
  (ICLR)}, 2016.

\bibitem[Li et~al.(2022)Li, Wu, Fan, Mangalam, Xiong, Malik, and
  Feichtenhofer]{li2022mvitv2}
Yanghao Li, Chao-Yuan Wu, Haoqi Fan, Karttikeya Mangalam, Bo Xiong, Jitendra
  Malik, and Christoph Feichtenhofer.
\newblock {MViTv2}: Improved multiscale vision transformers for classification
  and detection.
\newblock In \emph{Proceedings of the IEEE/CVF Conference on Computer Vision
  and Pattern Recognition (CVPR)}, pages 4804--4814, 2022.

\bibitem[Lin et~al.(2014)Lin, Maire, Belongie, Hays, Perona, Ramanan,
  Doll{\'a}r, and Zitnick]{lin2014microsoft}
Tsung-Yi Lin, Michael Maire, Serge Belongie, James Hays, Pietro Perona, Deva
  Ramanan, Piotr Doll{\'a}r, and C.~Lawrence Zitnick.
\newblock Microsoft {COCO}: Common objects in context.
\newblock In \emph{European Conference on Computer Vision (ECCV)}, pages
  740--755. Springer, 2014.

\bibitem[Liu et~al.(2022)Liu, Ning, Cao, Wei, Zhang, Lin, and Hu]{liu2022video}
Ze Liu, Jia Ning, Yue Cao, Yixuan Wei, Zheng Zhang, Stephen Lin, and Han Hu.
\newblock Video swin transformer.
\newblock In \emph{Proceedings of the IEEE/CVF Conference on Computer Vision
  and Pattern Recognition (CVPR)}, pages 3202--3211, 2022.

\bibitem[Maiorca et~al.(2023)Maiorca, Moschella, Norelli, Fumero, Locatello,
  and Rodol{\`a}]{maiorca2023latent}
Valentino Maiorca, Luca Moschella, Antonio Norelli, Marco Fumero, Francesco
  Locatello, and Emanuele Rodol{\`a}.
\newblock Latent space translation via semantic alignment.
\newblock In \emph{Advances in Neural Information Processing Systems
  (NeurIPS)}, 2023.

\bibitem[Maniparambil et~al.(2024)Maniparambil, Akshulakov, Dahou~Djilali,
  Narayan, Seddik, Mangalam, and O'Connor]{maniparambil2024vision}
Mayug Maniparambil, Raiymbek Akshulakov, Yasser~Abdelaziz Dahou~Djilali, Sanath
  Narayan, Mohamed El~Amine Seddik, Karttikeya Mangalam, and Noel~E. O'Connor.
\newblock Do vision and language encoders represent the world similarly?
\newblock In \emph{Proceedings of the IEEE/CVF Conference on Computer Vision
  and Pattern Recognition (CVPR)}, 2024.

\bibitem[Mashour et~al.(2020)Mashour, Roelfsema, Changeux, and
  Dehaene]{mashour}
George~A. Mashour, Pieter Roelfsema, Jean-Pierre Changeux, and Stanislas
  Dehaene.
\newblock Conscious processing and the global neuronal workspace hypothesis.
\newblock \emph{Neuron}, 105\penalty0 (5):\penalty0 776--798, 2020.

\bibitem[McCann(2026)]{mccann2026polymorphism}
Jordan~F. McCann.
\newblock Polymorphism is rotation: Operational mechanistic interpretability
  from a two-layer transformer to {Pythia-70m}.
\newblock \emph{arXiv preprint arXiv:2605.24577}, 2026.

\bibitem[Merity et~al.(2017)Merity, Xiong, Bradbury, and
  Socher]{merity2017pointer}
Stephen Merity, Caiming Xiong, James Bradbury, and Richard Socher.
\newblock Pointer sentinel mixture models.
\newblock In \emph{International Conference on Learning Representations
  (ICLR)}, 2017.

\bibitem[Merullo et~al.(2023)Merullo, Castricato, Eickhoff, and
  Pavlick]{merullo2023linearly}
Jack Merullo, Louis Castricato, Carsten Eickhoff, and Ellie Pavlick.
\newblock Linearly mapping from image to text space.
\newblock In \emph{International Conference on Learning Representations
  (ICLR)}, 2023.

\bibitem[Mikolov et~al.(2013)Mikolov, Le, and Sutskever]{mikolov2013exploiting}
Tomas Mikolov, Quoc~V. Le, and Ilya Sutskever.
\newblock Exploiting similarities among languages for machine translation.
\newblock \emph{arXiv preprint arXiv:1309.4168}, 2013.

\bibitem[Miller(1995)]{miller1995wordnet}
George~A. Miller.
\newblock {WordNet}: A lexical database for {English}.
\newblock \emph{Communications of the ACM}, 38\penalty0 (11):\penalty0 39--41,
  1995.

\bibitem[Moayeri et~al.(2023)Moayeri, Rezaei, Sanjabi, and
  Feizi]{moayeri2023text}
Mazda Moayeri, Keivan Rezaei, Maziar Sanjabi, and Soheil Feizi.
\newblock Text-to-concept (and back) via cross-model alignment.
\newblock In \emph{Proceedings of the 40th International Conference on Machine
  Learning (ICML)}, pages 25037--25060, 2023.

\bibitem[Morcos et~al.(2018)Morcos, Raghu, and Bengio]{morcos2018insights}
Ari~S. Morcos, Maithra Raghu, and Samy Bengio.
\newblock Insights on representational similarity in neural networks with
  canonical correlation.
\newblock In \emph{Advances in Neural Information Processing Systems
  (NeurIPS)}, 2018.

\bibitem[Moschella et~al.(2023)Moschella, Maiorca, Fumero, Norelli, Locatello,
  and Rodolà]{moschella2023relative}
Luca Moschella, Valentino Maiorca, Marco Fumero, Antonio Norelli, Francesco
  Locatello, and Emanuele Rodolà.
\newblock Relative representations enable zero-shot latent space communication.
\newblock In \emph{International Conference on Learning Representations
  (ICLR)}, 2023.

\bibitem[Neo et~al.(2025)Neo, Ong, Torr, Geva, Krueger, and
  Barez]{neo2025towards}
Clement Neo, Luke Ong, Philip Torr, Mor Geva, David Krueger, and Fazl Barez.
\newblock Towards interpreting visual information processing in vision-language
  models.
\newblock In \emph{International Conference on Learning Representations
  (ICLR)}, 2025.

\bibitem[Norouzi et~al.(2014)Norouzi, Mikolov, Bengio, Singer, Shlens, Frome,
  Corrado, and Dean]{norouzi2014zero}
Mohammad Norouzi, Tomas Mikolov, Samy Bengio, Yoram Singer, Jonathon Shlens,
  Andrea Frome, Greg~S. Corrado, and Jeffrey Dean.
\newblock Zero-shot learning by convex combination of semantic embeddings.
\newblock In \emph{International Conference on Learning Representations
  (ICLR)}, 2014.

\bibitem[{Nostalgebraist}(2020)]{nostalgebraist2020interpreting}
{Nostalgebraist}.
\newblock Interpreting {GPT}: The logit lens.
\newblock LessWrong, \url{https://www.lesswrong.com/posts/AcKRB8wDpdaN6v6ru},
  2020.
\newblock Posted 31 Aug 2020. Accessed 2026-10-31.

\bibitem[Oozeer et~al.(2025)Oozeer, Nathawani, Prakash, Lan, Harrasse, and
  Abdullah]{oozeer2025activation}
Narmeen~Fatimah Oozeer, Dhruv Nathawani, Nirmalendu Prakash, Michael Lan, Abir
  Harrasse, and Amir Abdullah.
\newblock Activation space interventions can be transferred between large
  language models.
\newblock In \emph{Proceedings of the 42nd International Conference on Machine
  Learning (ICML)}, pages 47212--47286, 2025.

\bibitem[Oquab et~al.(2024)Oquab, Darcet, Moutakanni, et~al.]{oquab2024dinov2}
Maxime Oquab, Timothée Darcet, Théo Moutakanni, et~al.
\newblock {DINOv2}: Learning robust visual features without supervision.
\newblock \emph{Transactions on Machine Learning Research}, 2024.

\bibitem[Park et~al.(2024)Park, Choe, and Veitch]{park2024linear}
Kiho Park, Yo~Joong Choe, and Victor Veitch.
\newblock The linear representation hypothesis and the geometry of large
  language models.
\newblock In \emph{Proceedings of the 41st International Conference on Machine
  Learning (ICML)}, pages 39643--39666, 2024.

\bibitem[Pennington et~al.(2014)Pennington, Socher, and
  Manning]{pennington2014glove}
Jeffrey Pennington, Richard Socher, and Christopher Manning.
\newblock {GloVe}: Global vectors for word representation.
\newblock In \emph{Proceedings of the 2014 Conference on Empirical Methods in
  Natural Language Processing ({EMNLP})}, pages 1532--1543, Doha, Qatar, 2014.
  Association for Computational Linguistics.

\bibitem[Plato(c.~375 BC)]{plato_cave}
Plato.
\newblock Republic, c.~375 BC.

\bibitem[Podell et~al.(2024)Podell, English, Lacey, Blattmann, Dockhorn,
  M{\"u}ller, Penna, and Rombach]{podell2024sdxl}
Dustin Podell, Zion English, Kyle Lacey, Andreas Blattmann, Tim Dockhorn, Jonas
  M{\"u}ller, Joe Penna, and Robin Rombach.
\newblock {SDXL}: Improving latent diffusion models for high-resolution image
  synthesis.
\newblock In \emph{International Conference on Learning Representations
  (ICLR)}, 2024.

\bibitem[Poppi et~al.(2026)Poppi, Cappelletti, Sarto, Schiffers, Kessler,
  Cornia, Baraldi, and Cucchiara]{poppi2026share}
Tobia Poppi, Silvia Cappelletti, Sara Sarto, Florian Schiffers, Garin Kessler,
  Marcella Cornia, Lorenzo Baraldi, and Rita Cucchiara.
\newblock Do models share safety representations? {C}ross-model steering for
  safe visual generation.
\newblock \emph{arXiv preprint arXiv:2606.05290}, 2026.

\bibitem[Puri et~al.(2026)Puri, Berend, Lapuschkin, and Samek]{puri2026atlas}
Bruno Puri, Jim Berend, Sebastian Lapuschkin, and Wojciech Samek.
\newblock Atlas-alignment: Making interpretability transferable across language
  models.
\newblock In \emph{ICLR 2026 Workshop on Representational Alignment
  (Re-Align)}, 2026.

\bibitem[Putnam(1982)]{putnam1982three}
Hilary Putnam.
\newblock Three kinds of scientific realism.
\newblock \emph{The Philosophical Quarterly}, 32\penalty0 (128):\penalty0
  195--200, 1982.

\bibitem[Radford et~al.(2021)Radford, Kim, Hallacy,
  et~al.]{radford2021learning}
Alec Radford, Jong~Wook Kim, Chris Hallacy, et~al.
\newblock Learning transferable visual models from natural language
  supervision.
\newblock In \emph{Proceedings of the 38th International Conference on Machine
  Learning (ICML)}, pages 8748--8763, 2021.

\bibitem[Reizinger et~al.(2025)Reizinger, Bizeul, Juhos, Vogt, Balestriero,
  Brendel, and Klindt]{reizinger2025cross}
Patrik Reizinger, Alice Bizeul, Attila Juhos, Julia~E. Vogt, Randall
  Balestriero, Wieland Brendel, and David Klindt.
\newblock Cross-entropy is all you need to invert the data generating process.
\newblock In \emph{International Conference on Learning Representations
  (ICLR)}, 2025.

\bibitem[Richens and Everitt(2024)]{richens2024robust}
Jonathan Richens and Tom Everitt.
\newblock Robust agents learn causal world models.
\newblock In \emph{International Conference on Learning Representations
  (ICLR)}, 2024.

\bibitem[Rimsky et~al.(2024)Rimsky, Gabrieli, Schulz, Tong, Hubinger, and
  Turner]{rimsky2024steering}
Nina Rimsky, Nick Gabrieli, Julian Schulz, Meg Tong, Evan Hubinger, and
  Alexander Turner.
\newblock Steering {Llama} 2 via contrastive activation addition.
\newblock In \emph{Proceedings of the 62nd Annual Meeting of the Association
  for Computational Linguistics (ACL)}, pages 15504--15522, 2024.

\bibitem[Roeder et~al.(2021)Roeder, Metz, and Kingma]{roeder2021linear}
Geoffrey Roeder, Luke Metz, and Diederik~P. Kingma.
\newblock On linear identifiability of learned representations.
\newblock In \emph{Proceedings of the 38th International Conference on Machine
  Learning (ICML)}, pages 9030--9039, 2021.

\bibitem[Schnaus et~al.(2025)Schnaus, Araslanov, and Cremers]{schnaus2025blind}
Dominik Schnaus, Nikita Araslanov, and Daniel Cremers.
\newblock It's a (blind) match! {T}owards vision-language correspondence
  without parallel data.
\newblock In \emph{Proceedings of the IEEE/CVF Conference on Computer Vision
  and Pattern Recognition (CVPR)}, 2025.

\bibitem[Schnitzer et~al.(2012)Schnitzer, Flexer, Schedl, and
  Widmer]{schnitzer2012local}
Dominik Schnitzer, Arthur Flexer, Markus Schedl, and Gerhard Widmer.
\newblock Local and global scaling reduce hubs in space.
\newblock \emph{Journal of Machine Learning Research}, 13:\penalty0 2871--2902,
  2012.

\bibitem[Sch{\"o}nemann(1966)]{schonemann1966generalized}
Peter~H. Sch{\"o}nemann.
\newblock A generalized solution of the orthogonal {Procrustes} problem.
\newblock \emph{Psychometrika}, 31\penalty0 (1):\penalty0 1--10, 1966.

\bibitem[Schuhmann et~al.(2022)Schuhmann, Beaumont, Vencu,
  et~al.]{schuhmann2022laion}
Christoph Schuhmann, Romain Beaumont, Richard Vencu, et~al.
\newblock {LAION-5B}: An open large-scale dataset for training next generation
  image-text models.
\newblock In \emph{Advances in Neural Information Processing Systems (NeurIPS),
  Datasets and Benchmarks Track}, 2022.

\bibitem[Sim{\'e}oni et~al.(2025)Sim{\'e}oni, Vo, Seitzer, Baldassarre, Oquab,
  et~al.]{simeoni2025dinov3}
Oriane Sim{\'e}oni, Huy~V. Vo, Maximilian Seitzer, Federico Baldassarre, Maxime
  Oquab, et~al.
\newblock {DINOv3}.
\newblock \emph{arXiv preprint arXiv:2508.10104}, 2025.

\bibitem[Smith et~al.(2025)Smith, Mannering, and Marcu]{smith2025functional}
Damian Smith, Harvey Mannering, and Antonia Marcu.
\newblock Functional alignment can mislead: Examining model stitching.
\newblock In \emph{Proceedings of the 42nd International Conference on Machine
  Learning (ICML)}, pages 55972--55998, 2025.

\bibitem[Socher et~al.(2013)Socher, Ganjoo, Manning, and Ng]{socher2013zero}
Richard Socher, Milind Ganjoo, Christopher~D. Manning, and Andrew~Y. Ng.
\newblock Zero-shot learning through cross-modal transfer.
\newblock In \emph{Advances in Neural Information Processing Systems
  (NeurIPS)}, pages 935--943, 2013.

\bibitem[Song et~al.(2012)Song, Smola, Gretton, Bedo, and
  Borgwardt]{song2012feature}
Le Song, Alex Smola, Arthur Gretton, Justin Bedo, and Karsten Borgwardt.
\newblock Feature selection via dependence maximization.
\newblock \emph{Journal of Machine Learning Research}, 13\penalty0
  (47):\penalty0 1393--1434, 2012.

\bibitem[Steiner et~al.(2022)Steiner, Kolesnikov, Zhai, Wightman, Uszkoreit,
  and Beyer]{steiner2022how}
Andreas Steiner, Alexander Kolesnikov, Xiaohua Zhai, Ross Wightman, Jakob
  Uszkoreit, and Lucas Beyer.
\newblock How to train your {ViT}? {Data}, augmentation, and regularization in
  vision transformers.
\newblock \emph{Transactions on Machine Learning Research}, 2022.

\bibitem[Tan et~al.(2024)Tan, Chanin, Lynch, Kanoulas, Paige, Garriga-Alonso,
  and Kirk]{tan2024analyzing}
Daniel Tan, David Chanin, Aengus Lynch, Dimitrios Kanoulas, Brooks Paige,
  Adri{\`a} Garriga-Alonso, and Robert Kirk.
\newblock Analyzing the generalization and reliability of steering vectors.
\newblock In \emph{Advances in Neural Information Processing Systems
  (NeurIPS)}, 2024.

\bibitem[{Team OLMo} et~al.(2024){Team OLMo}, Walsh, Soldaini,
  et~al.]{teamolmo2024olmo}
{Team OLMo}, Pete Walsh, Luca Soldaini, et~al.
\newblock 2 {OLMo} 2 furious.
\newblock \emph{arXiv preprint arXiv:2501.00656}, 2024.
\newblock A shorter version appeared at COLM 2025.

\bibitem[Theus et~al.(2025)Theus, Cabodi, Anagnostidis, Orvieto, Singh, and
  Boeva]{theus2025generalized}
Alexander Theus, Alessandro Cabodi, Sotiris Anagnostidis, Antonio Orvieto,
  Sidak~Pal Singh, and Valentina Boeva.
\newblock Generalized linear mode connectivity for transformers.
\newblock In \emph{Advances in Neural Information Processing Systems
  (NeurIPS)}, 2025.

\bibitem[Tian et~al.(2020)Tian, Krishnan, and Isola]{tian2020contrastive}
Yonglong Tian, Dilip Krishnan, and Phillip Isola.
\newblock Contrastive multiview coding.
\newblock In \emph{European Conference on Computer Vision (ECCV)}, pages
  776--794, 2020.

\bibitem[Timkey and van Schijndel(2021)]{timkey2021all}
William Timkey and Marten van Schijndel.
\newblock All bark and no bite: Rogue dimensions in transformer language models
  obscure representational quality.
\newblock In \emph{Proceedings of the 2021 Conference on Empirical Methods in
  Natural Language Processing (EMNLP)}, 2021.

\bibitem[Toker et~al.(2024)Toker, Orgad, Ventura, Arad, and
  Belinkov]{toker2024diffusion}
Michael Toker, Hadas Orgad, Mor Ventura, Dana Arad, and Yonatan Belinkov.
\newblock Diffusion lens: Interpreting text encoders in text-to-image
  pipelines.
\newblock In \emph{Proceedings of the 62nd Annual Meeting of the Association
  for Computational Linguistics (Volume 1: Long Papers)}, pages 9713--9728,
  Bangkok, Thailand, 2024. Association for Computational Linguistics.

\bibitem[Tong et~al.(2022)Tong, Song, Wang, and Wang]{tong2022videomae}
Zhan Tong, Yibing Song, Jue Wang, and Limin Wang.
\newblock {VideoMAE}: Masked autoencoders are data-efficient learners for
  self-supervised video pre-training.
\newblock In \emph{Advances in Neural Information Processing Systems
  (NeurIPS)}, 2022.

\bibitem[Touvron et~al.(2021)Touvron, Cord, Douze, Massa, Sablayrolles, and
  J{\'e}gou]{touvron2021training}
Hugo Touvron, Matthieu Cord, Matthijs Douze, Francisco Massa, Alexandre
  Sablayrolles, and Herv{\'e} J{\'e}gou.
\newblock Training data-efficient image transformers \& distillation through
  attention.
\newblock In \emph{Proceedings of the 38th International Conference on Machine
  Learning (ICML)}, 2021.

\bibitem[Touvron et~al.(2022)Touvron, Cord, and J{\'e}gou]{touvron2022deit}
Hugo Touvron, Matthieu Cord, and Herv{\'e} J{\'e}gou.
\newblock {DeiT III}: Revenge of the {ViT}.
\newblock In \emph{European Conference on Computer Vision (ECCV)}, 2022.

\bibitem[Tran et~al.(2018)Tran, Wang, Torresani, Ray, LeCun, and
  Paluri]{tran2018closer}
Du Tran, Heng Wang, Lorenzo Torresani, Jamie Ray, Yann LeCun, and Manohar
  Paluri.
\newblock A closer look at spatiotemporal convolutions for action recognition.
\newblock In \emph{Proceedings of the IEEE/CVF Conference on Computer Vision
  and Pattern Recognition (CVPR)}, pages 6450--6459, 2018.

\bibitem[Tschannen et~al.(2025)Tschannen, Gritsenko, Wang,
  et~al.]{tschannen2025siglip}
Michael Tschannen, Alexey Gritsenko, Xiao Wang, et~al.
\newblock {SigLIP} 2: Multilingual vision-language encoders with improved
  semantic understanding, localization, and dense features.
\newblock \emph{arXiv preprint arXiv:2502.14786}, 2025.

\bibitem[Turner et~al.(2023)Turner, Thiergart, Leech, Udell, Vazquez, Mini, and
  MacDiarmid]{turner2023steering}
Alexander~Matt Turner, Lisa Thiergart, Gavin Leech, David Udell, Juan~J.
  Vazquez, Ulisse Mini, and Monte MacDiarmid.
\newblock Steering language models with activation engineering.
\newblock \emph{arXiv preprint arXiv:2308.10248}, 2023.

\bibitem[Wang et~al.(2026{\natexlab{a}})Wang, Isola, and Cheung]{wang2026truth}
Sophie~L. Wang, Phillip Isola, and Brian Cheung.
\newblock The truth lies somewhere in the middle (of the generated tokens).
\newblock In \emph{Proceedings of the 43rd International Conference on Machine
  Learning (ICML)}, 2026{\natexlab{a}}.

\bibitem[Wang et~al.(2026{\natexlab{b}})Wang, Wang, Seneviratne, Chen, Xia,
  Wang, and Halgamuge]{wang2026jcomm}
Xiangwei Wang, Wei Wang, Sachith Seneviratne, Ken Chen, Yu Xia, Peng Wang, and
  Saman Halgamuge.
\newblock {J-Comm}: {Jacobian}-guided latent communication between {LLM}
  agents.
\newblock \emph{arXiv preprint arXiv:2610.32871}, 2026{\natexlab{b}}.

\bibitem[Wang et~al.(2024)Wang, Siskind, and Wang]{wang2024great}
Yipei Wang, Jeffrey Siskind, and Xiaoqian Wang.
\newblock Great minds think alike: The universal convergence trend of input
  salience.
\newblock In \emph{Advances in Neural Information Processing Systems
  (NeurIPS)}, pages 71672--71704, 2024.

\bibitem[Williams et~al.(2021)Williams, Kunz, Kornblith, and
  Linderman]{williams2021generalized}
Alex~H. Williams, Erin Kunz, Simon Kornblith, and Scott~W. Linderman.
\newblock Generalized shape metrics on neural representations.
\newblock In \emph{Advances in Neural Information Processing Systems
  (NeurIPS)}, 2021.

\bibitem[Xian et~al.(2019)Xian, Lampert, Schiele, and Akata]{xian2019zero}
Yongqin Xian, Christoph~H. Lampert, Bernt Schiele, and Zeynep Akata.
\newblock Zero-shot learning---a comprehensive evaluation of the good, the bad
  and the ugly.
\newblock \emph{IEEE Transactions on Pattern Analysis and Machine
  Intelligence}, 41\penalty0 (9):\penalty0 2251--2265, 2019.

\bibitem[Xing et~al.(2015)Xing, Wang, Liu, and Lin]{xing2015normalized}
Chao Xing, Dong Wang, Chao Liu, and Yiye Lin.
\newblock Normalized word embedding and orthogonal transform for bilingual word
  translation.
\newblock In \emph{Proceedings of the 2015 Conference of the North American
  Chapter of the Association for Computational Linguistics: Human Language
  Technologies (NAACL-HLT)}, pages 1006--1011, 2015.

\bibitem[Yang et~al.(2025)Yang, Li, Yang, et~al.]{yang2025qwen3}
An Yang, Anfeng Li, Baosong Yang, et~al.
\newblock {Qwen3} technical report.
\newblock \emph{arXiv preprint arXiv:2505.09388}, 2025.

\bibitem[Yavari and Zamani~Esfahlani(2026)]{yavari2026beyond}
Amirhossein Yavari and Farnaz Zamani~Esfahlani.
\newblock Beyond activation alignment: The geometry of neural sensitivity.
\newblock \emph{arXiv preprint arXiv:2605.03222}, 2026.

\bibitem[Zhu et~al.(2025)Zhu, Chen, He, LeCun, and Liu]{zhu2025transformers}
Jiachen Zhu, Xinlei Chen, Kaiming He, Yann LeCun, and Zhuang Liu.
\newblock Transformers without normalization.
\newblock In \emph{Proceedings of the IEEE/CVF Conference on Computer Vision
  and Pattern Recognition (CVPR)}, pages 14901--14911, 2025.

\bibitem[Zimmermann et~al.(2021)Zimmermann, Sharma, Schneider, Bethge, and
  Brendel]{zimmermann2021contrastive}
Roland~S. Zimmermann, Yash Sharma, Steffen Schneider, Matthias Bethge, and
  Wieland Brendel.
\newblock Contrastive learning inverts the data generating process.
\newblock In \emph{Proceedings of the 38th International Conference on Machine
  Learning (ICML)}, 2021.

\end{thebibliography}
